%% file: main.tex
\documentclass[pdflatex,sn-nature]{sn-jnl}

\usepackage{graphicx}
\usepackage{multirow}
\usepackage{amsmath,amssymb,amsfonts}
\usepackage{textcomp}
\usepackage{booktabs}
\usepackage{array}
\usepackage{longtable}
\usepackage{adjustbox}
\usepackage[table]{xcolor}
\usepackage{placeins}
\usepackage{needspace}
\usepackage{microtype} 

\definecolor{TableSection}{gray}{0.85}
\definecolor{TableHeader}{gray}{0.85}
\definecolor{TableRowBlue}{gray}{0.975}
\definecolor{TableRowGrey}{gray}{0.93}
\definecolor{TableARIVE}{gray}{0.93}
\definecolor{TableRule}{gray}{0}

\arrayrulecolor{TableRule}

\makeatletter
\long\def\LT@makecaption#1#2#3{%
  \LT@mcol\LT@cols c{\hbox to\z@{\hss
    \parbox[t]\LTcapwidth{%
      \tablecaptionfont\raggedright
      #1{{\bfseries #2}\hskip2mm}#3\vphantom{y}\par
      \vskip\belowcaptionskip}%
    \hss}}}
\makeatother

\newlength{\RRGMainResultsWidth}
\newlength{\RRGTrainingAblationWidth}
\newlength{\RRGNLGFixedWidth}
\newlength{\RRGNLGLayoutWidth}

\makeatletter
\newcommand{\metricboldchar}[2]{%
  \begingroup
    \setbox\z@=\hbox{\mdseries #2}
    \hbox to \wd\z@{\hss\bfseries #1\hss}%
  \endgroup
}
\newcommand{\metricbold}[1]{\metricbold@split#1\@nil}
\def\metricbold@split#1.#2#3#4\@nil{%
  \metricboldchar{#1}{0}%
  \metricboldchar{.}{.}%
  \metricboldchar{#2}{0}%
  \metricboldchar{#3}{0}%
  \metricboldchar{#4}{0}%
}
\makeatother

\newcommand{\sigmark}[1]{%
  \makebox[0pt][l]{\kern0.08em\textsuperscript{#1}}%
}
\newcommand{\metricci}[3]{%
  \begingroup
    \renewcommand{\arraystretch}{0.78}%
    \begin{tabular}[c]{@{}c@{}}
      \begingroup
        \let\textbf\metricbold
        \makebox[5.4em][c]{#1}%
      \endgroup\\[-0.35ex]
      {\footnotesize\makebox[6.6em][c]{(#2, #3)}}%
    \end{tabular}%
  \endgroup
}


\renewcommand{\floatpagefraction}{0.90}

\renewcommand{\dblfloatpagefraction}{0.90}
\begin{document}

\title[ARIVE for explainable radiology report generation]{Interleaving anatomical inspection and diagnostic reasoning for explainable radiology report generation}

\author[1]{\fnm{Chentao} \sur{Huang}}\email{cthuang@mail.dlut.edu.cn}

\author[2]{\fnm{Runlong} \sur{Lin}}\email{longjunji1129@126.com}

\author[1]{\fnm{Zhijun} \sur{Wang}}\email{wzj@mail.dlut.edu.cn}

\author[1]{\fnm{Yunxiang} \sur{Li}}\email{lyxdlut@mail.dlut.edu.cn}

\author[1]{\fnm{Yuanyuan} \sur{Sun}}\email{syuan@dlut.edu.cn}

\author[1]{\fnm{Jian} \sur{Wang}}\email{wangjian@dlut.edu.cn}

\author[1]{\fnm{Hongfei} \sur{Lin}}\email{hflin@dlut.edu.cn}

\author*[1]{\fnm{Ling} \sur{Luo}}\email{lingluo@dlut.edu.cn}

\affil*[1]{\orgdiv{School of Computer Science and Technology}, \orgname{Dalian University of Technology}, \orgaddress{\street{No. 2 Linggong Road, Ganjingzi District}, \city{Dalian}, \postcode{116024}, \state{Liaoning}, \country{China}}}

\affil[2]{\orgdiv{Department of Nuclear Medicine}, \orgname{The Second Hospital of Dalian Medical University}, \orgaddress{\street{No. 467 Zhongshan Road, Shahekou District}, \city{Dalian}, \postcode{116023}, \state{Liaoning}, \country{China}}}

\abstract{Automated radiology report generation can draft clinically plausible reports, yet the pathway from visual evidence to reported findings remains opaque, leaving the diagnostic basis of these reports unexplained and the generated reports difficult to review or revise. We introduce ARIVE (Anatomy-guided Reasoning with Interleaved Visual Evidence), an explainable report generation framework that interleaves anatomical inspection with diagnostic reasoning prior to final report composition. To support this process, we constructed 159,416 multimodal chain-of-thought (MCoT) trajectories from MIMIC-CXR, wherein visual evidence from inspected anatomical regions informs subsequent clinical interpretation. Trained on these trajectories and further optimized using anatomical and clinical-content rewards, ARIVE achieved superior clinical-efficacy performance compared with strong baselines on MIMIC-CXR and the abnormality-centered MIMIC-ABN. ARIVE's reports were competitive with those of leading models in blinded expert grading, and its MCoT trajectories earned high ratings for diagnostic traceability. Furthermore, a paired review evaluation suggests that access to MCoT can aid in identifying errors within generated reports.}


\maketitle

\section{Introduction}
\label{sec:introduction}
Medical imaging reports translate visual observations into findings that inform diagnosis, treatment and follow-up. In automated medical report generation, fluent and clinically plausible language alone does not reveal how the image supports each finding. This lack of transparency is clinically consequential, as expert evaluations have shown that automated reports may contain substantial errors, including misidentified or incorrectly described findings~\cite{tanno2025collaboration}. Moreover, evaluations show that even when multimodal models answer correctly, their image interpretation and reasoning can be flawed~\cite{jin2024hidden}. Explainability should therefore be treated as a core requirement of automated reporting rather than an optional capability. Specifically, an explainable system should make the visual evidence for each finding inspectable and the diagnostic reasoning underlying it traceable, helping clinicians assess the report and identify unsupported content~\cite{chen2022explainable,kim2026transparency}.

Automated medical report generation has expanded beyond the image-to-text captioning~\cite{chen2020generating,zambrano2025clinically,li2025report} to address a broader range of clinical needs. Recent advances span multiple medical imaging domains~\cite{chen2025deep,park2026robust,liu2025multimodal,wu2026universal}, incorporate longitudinal imaging context~\cite{liu2026priorrg,zhang2025libra}, and support clinician-assisted reporting workflows~\cite{tanno2025collaboration}. These developments broaden the potential clinical applications of automated reporting, but do not necessarily clarify how image evidence supports the generated findings.

Several approaches make the information used in report generation more explicit. Attention visualizations highlight image regions associated with generated findings~\cite{zambrano2025clinically}, retrieval-based approaches present clinically similar cases~\cite{hou2025radar}, and graph-based methods use explicit clinical entities and relations to guide report generation~\cite{li2022cross,li2023dynamic,huang2025ddgip}. Although informative, these approaches do not reveal how evidence gathered from specific regions updates diagnostic interpretation and ultimately shapes the generated report. Chain-of-thought (CoT) reasoning~\cite{wei2022chain,liu2024medcot,mei2026fir} takes a different route by presenting intermediate diagnostic steps as a readable narrative. However, a textual CoT trajectory does not ensure that each reasoning step is grounded in the corresponding visual evidence. As the reasoning trajectory lengthens, later steps may rely increasingly on preceding text rather than the image, a pattern associated with visual forgetting~\cite{sun2025mitigating,cheng2026visual}. Such drift may allow unsupported assumptions to accumulate and propagate through the reasoning process, increasing the risk of hallucinated findings. CoT can therefore make a stated reasoning sequence explicit while leaving unclear which regional visual evidence supports the resulting findings.
Grounded report generation provides a complementary form of transparency by explicitly linking report findings to spatially localized image evidence. MAIRA-2~\cite{bannur2024maira} pairs findings with bounding boxes, UniBiomed~\cite{wu2026universal} connects generated descriptions with segmentation outputs, and ophthalmic vision-language segmentation systems~\cite{jin2026grounded} use lesion masks for verification. These spatial outputs improve report inspectability by indicating the location associated with a finding and providing evidence against which it can be assessed. However, elucidating how regional visual evidence informs the diagnostic reasoning underlying the final report remains an unresolved challenge.

Within this broader landscape, we focus on radiology report generation (RRG) for chest radiography, the most commonly performed medical imaging examination. We introduce ARIVE (Anatomy-guided Reasoning with Interleaved Visual Evidence), a framework for explainable chest X-ray (CXR) report generation. ARIVE interleaves anatomical inspection with diagnostic interpretation, using the visual evidence obtained at each inspection step to guide subsequent reasoning and ultimately compose the final report. To support this workflow, we constructed 159,416 report-derived multimodal chain-of-thought (MCoT) training trajectories from MIMIC-CXR~\cite{johnson2019mimic}, grounded in 29 Chest ImaGenome anatomical regions~\cite{wu2021chest}. We further introduce Anatomical and Clinical-Content Optimization (ACCO) to jointly optimize anatomical grounding, intermediate clinical content and final-report fidelity. Adding ACCO improved the measured clinical-content accuracy of both intermediate reasoning and final reports. 
Evaluated on MIMIC-CXR~\cite{johnson2019mimic} and the abnormality-centered MIMIC-ABN~\cite{ni2020learning}, ARIVE achieved the strongest overall clinical-efficacy performance among the compared models. Blinded grading showed that ARIVE maintained competitive report quality, while assessment of its MCoT trajectories demonstrated high diagnostic traceability, linking final-report claims to intermediate interpretations. Paired review further showed that additional erroneous reports were identified when the corresponding MCoT trajectories were available, although review time increased. We also release scripts covering MCoT dataset construction, model training and inference to support reproduction of the full ARIVE pipeline. Together, these findings support integrating regional inspection into report generation to make the links between visual evidence, intermediate interpretations and reported findings inspectable. This approach offers a transferable paradigm for clinically faithful and explainable report generation across medical imaging.

\section{Results}
\label{sec:results}

\subsection{Overview of ARIVE}
\label{sec:study_design}
\begin{figure*}[!t]
\centering
\includegraphics[width=\textwidth]{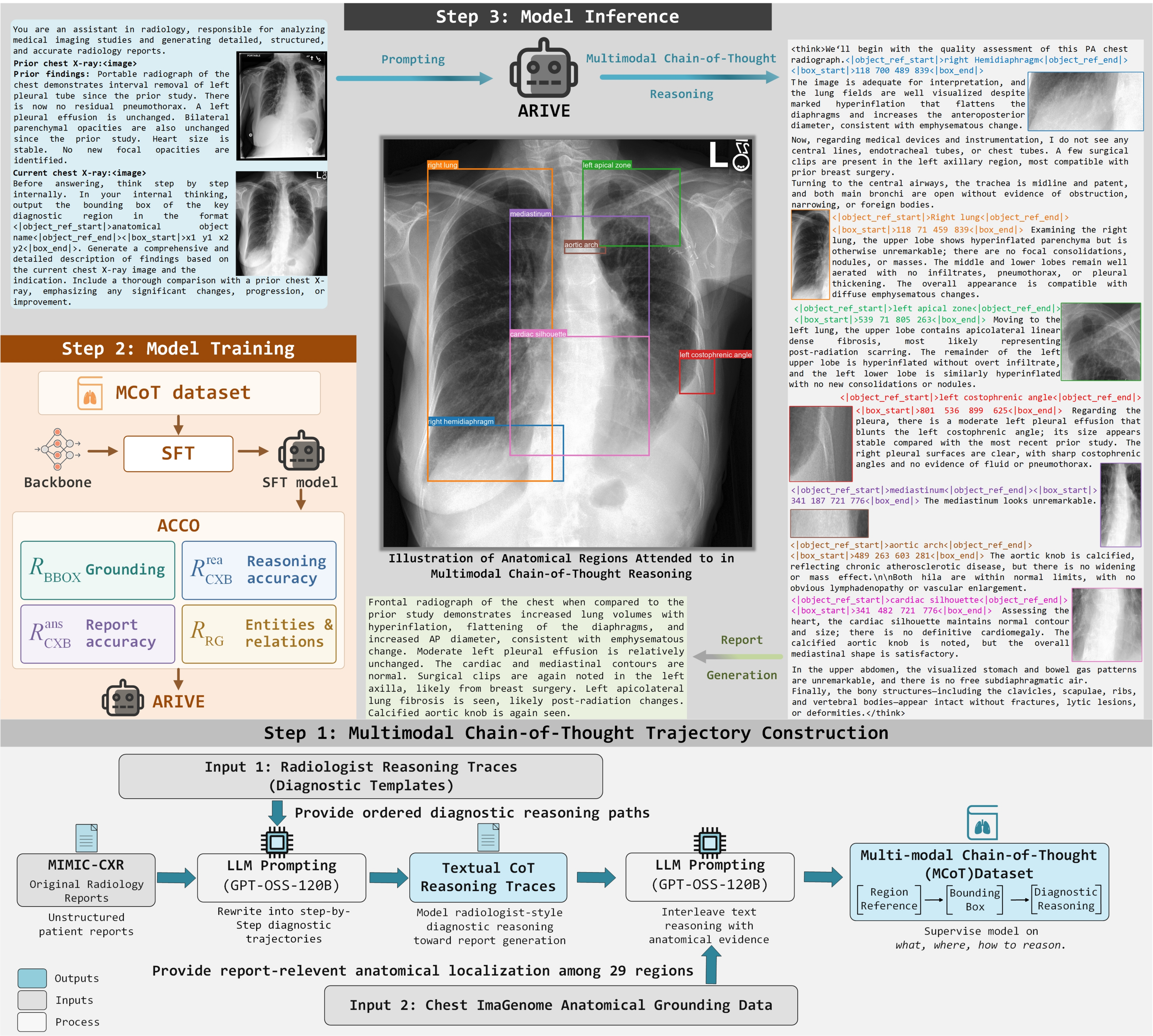}
\caption{ARIVE workflow for dataset construction, model training and report generation. Step 1 (bottom), radiologist-style diagnostic templates guide GPT-OSS-120B in converting MIMIC-CXR reports into textual reasoning trajectories. Chest ImaGenome annotations ground report-relevant steps in 29 anatomical regions to construct the multimodal chain-of-thought (MCoT) dataset. Step 2 (left), supervised fine-tuning (SFT) is followed by Anatomical and Clinical-Content Optimization (ACCO). Four rewards assess anatomical grounding, observation accuracy in intermediate reasoning and final reports, and clinical entity--relation agreement with reference reports. Step 3 (top and right), ARIVE interleaves anatomical inspection with diagnostic reasoning before generating the final report. CXB and RG denote CheXbert and RadGraph, respectively.}
\label{fig:workflow}
\end{figure*}

ARIVE follows a three-step workflow that links MCoT trajectory construction, model training and MCoT-guided model inference (Fig.~\ref{fig:workflow}). First, we reorganized MIMIC-CXR reports into 159,416 diagnostic trajectories using radiologist-style templates~\cite{myronenko2025reasoning} and grounded clinically significant steps in 29 Chest ImaGenome anatomical regions~\cite{wu2021chest}. Second, we trained ARIVE with supervised fine-tuning (SFT) to interleave anatomical inspection with regional interpretation before composing the final report, followed by optimization with ACCO. ACCO combines four rewards for anatomical grounding ($R_{\mathrm{BBOX}}$), observation-level accuracy in the intermediate reasoning ($R^{\mathrm{rea}}_{\mathrm{CXB}}$) and the final report ($R^{\mathrm{ans}}_{\mathrm{CXB}}$), and preservation of the clinical entities and relations in the reference report ($R_{\mathrm{RG}}$). For the grounding and CheXbert rewards, inverse-frequency weighting assigns greater weight to underrepresented anatomical regions and clinical observations. Third, during inference, ARIVE first surveys the radiograph globally and then iteratively inspects diagnostically relevant anatomical regions, using the resulting visual evidence to guide its ongoing interpretation and, ultimately, the final report.

We trained and evaluated dataset-specific ARIVE models on MIMIC-CXR, the primary benchmark, and on MIMIC-ABN. MIMIC-ABN comprises abnormal studies only, with reports restricted to abnormality-related findings. This abnormality-centered setting provides a focused assessment of abnormal-finding reporting while reducing the influence of normal-report content. Evaluation focused on clinical efficacy metrics, measured with CheXbert~\cite{smit2020combining}, $\mathrm{SRR}\text{-}\mathrm{BERT}$~\cite{delbrouck2025automated} and RadGraph~\cite{jain2021radgraph} on both datasets and additionally with GREEN~\cite{ostmeier2024green} on MIMIC-CXR. Lexical similarity to the reference reports was assessed with three natural language generation (NLG) metrics: BLEU-4~\cite{papineni2002bleu}, METEOR~\cite{banerjee2005meteor} and ROUGE-L~\cite{lin2004rouge}. Anatomical grounding was evaluated on the corresponding Chest ImaGenome Gold subsets using region-selection $\mathrm{F}_{1}$ and mean intersection over union (mIoU).

\subsection{Comparison with existing report-generation models}
\label{sec:comparison}
\begin{figure*}[!p]
\centering
\includegraphics[width=\textwidth,height=0.85\textheight,keepaspectratio]{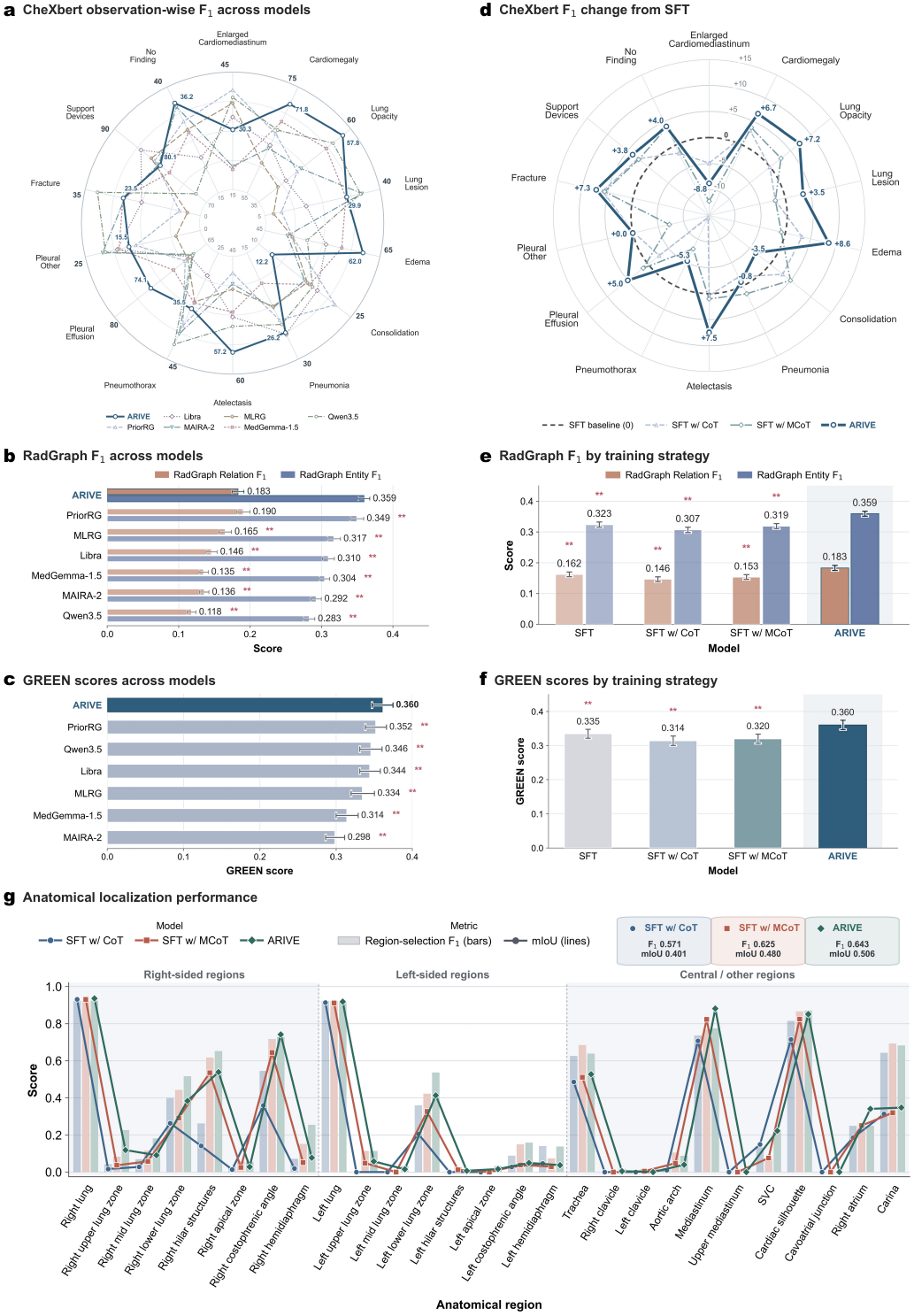}
\caption{Clinical efficacy and anatomical grounding on MIMIC-CXR. (a)--(c), Comparison of ARIVE and six comparators using observation-wise CheXbert $\mathrm{F}_{1}$ (a), RadGraph relation and entity $\mathrm{F}_{1}$ (b), and GREEN (c). In (a), inner and outer tick labels show observation-specific $\mathrm{F}_{1}$ axis limits (polygon area does not indicate overall performance), and annotations give ARIVE's scores. (d)--(f), Comparison of SFT, SFT w/ CoT, SFT w/ MCoT and ARIVE using CheXbert $\mathrm{F}_{1}$ changes from SFT in percentage points (d), RadGraph $\mathrm{F}_{1}$ (e), and GREEN (f). (g), Region-selection $\mathrm{F}_{1}$ (bars) and mean intersection over union (mIoU; lines) across all ground-truth regions in 636 Chest ImaGenome Gold examples. Error bars in (b,c,e,f) denote 95\% confidence intervals. Asterisks indicate significantly higher scores for ARIVE in comparisons with the indicated models or training settings: $^{*}p < 0.05$; $^{**}p < 0.01$.}
\label{fig:cxr_evaluation_composite}
\end{figure*}

We first compared ARIVE on MIMIC-CXR with six models from three categories: a general-purpose multimodal large language model (MLLM), Qwen3.5-397B-A17B~\cite{qwen3.5} (hereafter Qwen3.5); a generalist medical MLLM, MedGemma-1.5~\cite{sellergren2026medgemma}; and four task-specific report-generation models, MAIRA-2~\cite{bannur2024maira}, Libra~\cite{zhang2025libra}, MLRG~\cite{liu2025enhanced} and PriorRG~\cite{liu2026priorrg}. ARIVE achieved the highest scores across all seven clinical-efficacy measures: micro- and macro-averaged CheXbert $\mathrm{F}_{1}$ over five common and all 14 observations, $\mathrm{F}_{1}\text{-}\mathrm{SRR}\text{-}\mathrm{BERT}$, RadGraph $\mathrm{F}_{1}$ and GREEN (Table~\ref{tab:main_results} and Fig.~\ref{fig:cxr_evaluation_composite}c). Its advantage was significant ($p < 0.01$) in 40 of the 42 comparisons. Both exceptions involved the strongest comparator on the respective measure: 14-observation macro $\mathrm{F}_{1}$ versus Qwen3.5 and $\mathrm{F}_{1}\text{-}\mathrm{SRR}\text{-}\mathrm{BERT}$ versus PriorRG.

The comparators showed complementary strengths. On the CheXbert measures, the two MLLMs were the strongest comparators. Qwen3.5 led on the primary 14-observation macro $\mathrm{F}_{1}$, on which ARIVE achieved 0.437 (95\% CI, 0.416--0.453) compared with 0.428 (0.407--0.444), and MedGemma-1.5 led on the other three, on which ARIVE scored 3.9--9.4\% higher (for example, 0.603 versus 0.556 for 14-observation micro $\mathrm{F}_{1}$). At the observation level, ARIVE achieved the highest CheXbert $\mathrm{F}_{1}$ for Cardiomegaly, Lung Opacity, Edema, Atelectasis, Pleural Effusion and No Finding (Fig.~\ref{fig:cxr_evaluation_composite}a).

The task-specific models generally showed greater agreement with reference reports in clinical entities and their relations: three of the four exceeded both MLLMs on RadGraph $\mathrm{F}_{1}$, and PriorRG led the comparators at 0.270. ARIVE scored slightly higher, at 0.271 (Table~\ref{tab:main_results} and Fig.~\ref{fig:cxr_evaluation_composite}b). GREEN complements these measures by weighing matched findings against clinically significant errors, including false or missed findings and errors in location, severity or comparison with prior studies~\cite{ostmeier2024green}. ARIVE achieved the highest GREEN score at 0.360, an increase of 2.3\% over  the second-ranked PriorRG (0.352; Fig.~\ref{fig:cxr_evaluation_composite}c). ARIVE also achieved the highest $\mathrm{F}_{1}\text{-}\mathrm{SRR}\text{-}\mathrm{BERT}$, which captures fine-grained findings across 55 radiology labels (0.515 versus 0.511 for PriorRG).

\begin{table*}[!t]
  \centering
  \caption{Clinical efficacy performance of ARIVE and comparator models on MIMIC-CXR and MIMIC-ABN. CheXbert $\mathrm{F}_{1}$ subscripts indicate micro (mic) or macro (mac) averaging over five common or all 14 clinical observations. $\mathrm{RG}\text{-}\mathrm{F}_{1}$ denotes RadGraph $\mathrm{F}_{1}$. Values are reported with 95\% CIs in parentheses. Bold marks the highest value for each metric within each dataset. Asterisks indicate significantly higher scores for ARIVE than for the indicated comparator in one-sided comparisons using patient-clustered paired permutation tests for CheXbert and SRR-BERT and paired Wilcoxon signed-rank tests for RadGraph. $^{*}p < 0.05$; $^{**}p < 0.01$.}
  \label{tab:main_results}
  \begingroup
  \renewcommand{\arraystretch}{1.80}
  \setlength{\tabcolsep}{2.5pt}
  \begin{adjustbox}{gstore width=\RRGMainResultsWidth,max width=\textwidth}
  \begin{tabular}{clcccccc}
    \toprule
    \rowcolor{TableHeader}
    \textbf{Dataset}
    & \textbf{Model}
    & $\mathrm{F}_{\mathrm{1,mic}\text{-}5}$
    & $\mathrm{F}_{\mathrm{1,mac}\text{-}5}$
    & $\mathrm{F}_{\mathrm{1,mic}\text{-}14}$
    & $\mathrm{F}_{\mathrm{1,mac}\text{-}14}$
    & $\mathrm{F}_{1}\text{-}\mathrm{SRR}\text{-}\mathrm{BERT}$
    & $\mathrm{RG}\text{-}\mathrm{F}_{1}$ \\
    \midrule
    \rowcolor{TableRowBlue}
    & Qwen3.5-397B-A17B & \metricci{0.567\sigmark{**}}{0.545}{0.588} & \metricci{0.519\sigmark{**}}{0.498}{0.538} & \metricci{0.535\sigmark{**}}{0.518}{0.551} & \metricci{0.428}{0.407}{0.444} & \metricci{0.498\sigmark{**}}{0.488}{0.509} & \metricci{0.200\sigmark{**}}{0.193}{0.207} \\
    \cmidrule[0.50pt]{2-8}
    \rowcolor{TableRowGrey}
    & MedGemma-1.5 & \metricci{0.596\sigmark{**}}{0.572}{0.618} & \metricci{0.534\sigmark{**}}{0.513}{0.553} & \metricci{0.556\sigmark{**}}{0.539}{0.573} & \metricci{0.409\sigmark{**}}{0.388}{0.425} & \metricci{0.491\sigmark{**}}{0.479}{0.502} & \metricci{0.219\sigmark{**}}{0.213}{0.226} \\
    \cmidrule[0.50pt]{2-8}
    \rowcolor{TableRowBlue}
    & MAIRA-2 & \metricci{0.563\sigmark{**}}{0.540}{0.583} & \metricci{0.491\sigmark{**}}{0.469}{0.511} & \metricci{0.526\sigmark{**}}{0.508}{0.541} & \metricci{0.407\sigmark{**}}{0.386}{0.425} & \metricci{0.491\sigmark{**}}{0.478}{0.502} & \metricci{0.214\sigmark{**}}{0.207}{0.221} \\
    \cmidrule[0.50pt]{2-8}
    \rowcolor{TableRowGrey}
    & Libra & \metricci{0.583\sigmark{**}}{0.557}{0.605} & \metricci{0.519\sigmark{**}}{0.496}{0.539} & \metricci{0.537\sigmark{**}}{0.516}{0.556} & \metricci{0.395\sigmark{**}}{0.374}{0.412} & \metricci{0.489\sigmark{**}}{0.474}{0.503} & \metricci{0.228\sigmark{**}}{0.220}{0.236} \\
    \cmidrule[0.50pt]{2-8}
    \rowcolor{TableRowBlue}
    & MLRG & \metricci{0.559\sigmark{**}}{0.535}{0.580} & \metricci{0.491\sigmark{**}}{0.468}{0.512} & \metricci{0.520\sigmark{**}}{0.502}{0.537} & \metricci{0.363\sigmark{**}}{0.343}{0.379} & \metricci{0.492\sigmark{**}}{0.479}{0.504} & \metricci{0.241\sigmark{**}}{0.232}{0.250} \\
    \cmidrule[0.50pt]{2-8}
    \rowcolor{TableRowGrey}
    & PriorRG & \metricci{0.575\sigmark{**}}{0.552}{0.597} & \metricci{0.508\sigmark{**}}{0.484}{0.531} & \metricci{0.537\sigmark{**}}{0.518}{0.555} & \metricci{0.391\sigmark{**}}{0.370}{0.408} & \metricci{0.511}{0.498}{0.523} & \metricci{0.270\sigmark{**}}{0.261}{0.279} \\
    \cmidrule[0.50pt]{2-8}
    \rowcolor{TableARIVE}
    \multirow{-7}{*}{MIMIC-CXR} & ARIVE (ours) & \metricci{\textbf{0.652}}{0.630}{0.672} & \metricci{\textbf{0.555}}{0.536}{0.571} & \metricci{\textbf{0.603}}{0.586}{0.617} & \metricci{\textbf{0.437}}{0.416}{0.453} & \metricci{\textbf{0.515}}{0.503}{0.526} & \metricci{\textbf{0.271}}{0.263}{0.279} \\
    \midrule
    \rowcolor{TableRowBlue}
    & Qwen3.5-397B-A17B & \metricci{0.486\sigmark{**}}{0.420}{0.545} & \metricci{0.451\sigmark{**}}{0.384}{0.508} & \metricci{0.459\sigmark{**}}{0.423}{0.491} & \metricci{0.365\sigmark{**}}{0.326}{0.392} & \metricci{0.434}{0.401}{0.470} & \metricci{0.165\sigmark{**}}{0.151}{0.177} \\
    \cmidrule[0.50pt]{2-8}
    \rowcolor{TableRowGrey}
    & MedGemma-1.5 & \metricci{0.529\sigmark{**}}{0.474}{0.581} & \metricci{0.479\sigmark{**}}{0.422}{0.529} & \metricci{0.481\sigmark{**}}{0.445}{0.517} & \metricci{0.360\sigmark{**}}{0.320}{0.392} & \metricci{0.421}{0.387}{0.459} & \metricci{0.165\sigmark{**}}{0.151}{0.178} \\
    \cmidrule[0.50pt]{2-8}
    \rowcolor{TableRowBlue}
    & MAIRA-2 & \metricci{0.522\sigmark{**}}{0.459}{0.577} & \metricci{0.469\sigmark{**}}{0.407}{0.522} & \metricci{0.466\sigmark{**}}{0.428}{0.504} & \metricci{0.360\sigmark{**}}{0.326}{0.385} & \metricci{0.383\sigmark{**}}{0.349}{0.420} & \metricci{0.164\sigmark{**}}{0.148}{0.180} \\
    \cmidrule[0.50pt]{2-8}
    \rowcolor{TableRowGrey}
    & Libra & \metricci{0.485\sigmark{**}}{0.439}{0.531} & \metricci{0.439\sigmark{**}}{0.392}{0.485} & \metricci{0.458\sigmark{**}}{0.423}{0.493} & \metricci{0.356\sigmark{**}}{0.322}{0.384} & \metricci{0.409\sigmark{**}}{0.376}{0.447} & \metricci{0.140\sigmark{**}}{0.127}{0.155} \\
    \cmidrule[0.50pt]{2-8}
    \rowcolor{TableRowBlue}
    & MLRG & \metricci{0.575}{0.525}{0.623} & \metricci{0.521}{0.462}{0.575} & \metricci{0.514\sigmark{*}}{0.474}{0.551} & \metricci{0.362\sigmark{*}}{0.333}{0.387} & \metricci{0.409\sigmark{**}}{0.373}{0.451} & \metricci{0.190\sigmark{**}}{0.168}{0.215} \\
    \cmidrule[0.50pt]{2-8}
    \rowcolor{TableRowGrey}
    & PriorRG & \metricci{0.514\sigmark{**}}{0.459}{0.564} & \metricci{0.456\sigmark{**}}{0.398}{0.507} & \metricci{0.476\sigmark{**}}{0.441}{0.510} & \metricci{0.332\sigmark{**}}{0.300}{0.361} & \metricci{0.404\sigmark{**}}{0.369}{0.441} & \metricci{0.191\sigmark{*}}{0.171}{0.211} \\
    \cmidrule[0.50pt]{2-8}
    \rowcolor{TableARIVE}
    \multirow{-7}{*}{MIMIC-ABN} & ARIVE (ours) & \metricci{\textbf{0.581}}{0.529}{0.631} & \metricci{\textbf{0.541}}{0.485}{0.590} & \metricci{\textbf{0.545}}{0.514}{0.574} & \metricci{\textbf{0.408}}{0.372}{0.433} & \metricci{\textbf{0.441}}{0.405}{0.477} & \metricci{\textbf{0.207}}{0.192}{0.222} \\
    \bottomrule
  \end{tabular}
  \end{adjustbox}
  \endgroup
\end{table*}

\label{sec:abnormality_enriched}
We next evaluated abnormality-centered report generation on MIMIC-ABN. ARIVE again achieved the highest scores across all six tabulated measures: the four CheXbert measures, $\mathrm{F}_{1}\text{-}\mathrm{SRR}\text{-}\mathrm{BERT}$ and RadGraph $\mathrm{F}_{1}$ (Table~\ref{tab:main_results}). On the primary 14-observation macro $\mathrm{F}_{1}$, ARIVE reached 0.408 (95\% CI, 0.372--0.433) compared with 0.365 (0.326--0.392) for Qwen3.5, the strongest comparator on this measure. The difference was larger than on MIMIC-CXR (0.043 versus 0.009) and was significant ($p < 0.01$). Overall, ARIVE's scores were significantly higher in 32 of the 36 comparisons ($p < 0.05$). The four exceptions were the two five-observation CheXbert measures against MLRG and F1-SRR-BERT against Qwen3.5 and MedGemma-1.5. These results also supported ARIVE's performance on automated clinical-content metrics in the abnormality-centered setting.

Lexical similarity showed a different pattern: ARIVE did not achieve the highest BLEU-4, METEOR or ROUGE-L score on either dataset (Supplementary Table~\ref{supp-tab:full_report_nlg_metrics}). These NLG metrics primarily assess overlap with reference wording and may miss clinically important errors in findings and their relations, which clinical-efficacy metrics assess more directly~\cite{liu19a,BABAR2021102075,YU2023100802}. This divergence is consistent with ARIVE's emphasis on clinical content rather than reference phrasing. We next performed staged ablations to assess the contributions of MCoT supervision and ACCO.

\subsection{Ablation studies}
\label{sec:training_strategy_ablation}

\subsubsection{Effect of MCoT supervision on report generation}
\label{sec:sft_training_analysis}

To assess the contribution of MCoT supervision, we compared report-only SFT with SFT on text-only CoT or MCoT trajectories (Table~\ref{tab:training_ablation}). On MIMIC-CXR, text-only CoT increased CheXbert $\mathrm{F}_{1}$ on both five-observation measures and on 14-observation micro $\mathrm{F}_{1}$, whereas the primary 14-observation macro $\mathrm{F}_{1}$ remained unchanged at 0.412. These gains did not extend to the complementary clinical-efficacy measures: $\mathrm{F}_{1}\text{-}\mathrm{SRR}\text{-}\mathrm{BERT}$, RadGraph $\mathrm{F}_{1}$ and GREEN all declined, with GREEN falling from 0.335 to 0.314 (Fig.~\ref{fig:cxr_evaluation_composite}e,f). MIMIC-ABN showed a similar pattern, with small increases in three of the four CheXbert measures accompanied by lower $\mathrm{F}_{1}\text{-}\mathrm{SRR}\text{-}\mathrm{BERT}$ and RadGraph $\mathrm{F}_{1}$. Thus, although text-only CoT made the intermediate diagnostic statements explicit and readable, it did not consistently improve the clinical efficacy of the resulting reports, and the decline in GREEN suggests that a textual rationale alone was insufficient to reduce clinically important errors.

To assess the contribution of regional visual evidence, we next compared text-only CoT with MCoT, which interleaves anatomical inspection with diagnostic interpretation. On MIMIC-CXR, MCoT increased scores on every clinical-efficacy measure, including GREEN (from 0.314 to 0.320; Fig.~\ref{fig:cxr_evaluation_composite}f), except the primary 14-observation macro $\mathrm{F}_{1}$, which decreased slightly from 0.412 to 0.410. On MIMIC-ABN, MCoT increased scores on all six tabulated measures (Table~\ref{tab:training_ablation}), and all three NLG metrics improved on both datasets (Supplementary Table~\ref{supp-tab:full_training_ablation_nlg}). Relative to text-only reasoning, interleaved regional evidence therefore improved the clinical efficacy of the generated reports, supporting the contribution of interleaved regional visual evidence to clinically faithful report generation.

On the 142-example MIMIC-ABN Chest ImaGenome Gold subset, MCoT also improved intermediate diagnostic reasoning and its consistency with the final report (Table~\ref{tab:beta_ablation}). Relative to text-only CoT, MCoT increased reasoning-stage macro $\mathrm{F}_{1}$ by 0.103 (0.413 versus 0.310) and final-report macro $\mathrm{F}_{1}$ by 0.037 (0.411 versus 0.374), and raised reasoning--report consistency macro $\mathrm{F}_{1}$ from 0.728 to 0.894. Interleaved regional evidence thus strengthened clinically relevant interpretation and helped findings propagate more consistently from intermediate reasoning to the final report.

\begin{table*}[!ht]
  \centering
  \caption{Clinical efficacy performance across training settings on MIMIC-CXR and MIMIC-ABN. CheXbert $\mathrm{F}_{1}$ subscripts indicate micro (mic) or macro (mac) averaging over five common or all 14 clinical observations. SFT uses report-only supervision; SFT w/ CoT and SFT w/ MCoT add text-only and region-interleaved diagnostic trajectories, respectively. ARIVE denotes SFT w/ MCoT followed by ACCO. Values are reported with 95\% CIs in parentheses. Bold marks the highest value for each metric within each dataset. Asterisks indicate significantly higher scores for ARIVE than for the indicated training setting, under the tests specified in Table~\ref{tab:main_results}. $^{*}p < 0.05$; $^{**}p < 0.01$.}
  \label{tab:training_ablation}
  \begingroup
  \renewcommand{\arraystretch}{1.75}
  \setlength{\tabcolsep}{2.5pt}
  \begin{adjustbox}{gstore width=\RRGTrainingAblationWidth,max width=\textwidth}
  \begin{tabular}{clcccccc}
    \toprule
    \rowcolor{TableHeader}
    \textbf{Dataset}
    & \textbf{Training setting}
    & $\mathrm{F}_{\mathrm{1,mic}\text{-}5}$
    & $\mathrm{F}_{\mathrm{1,mac}\text{-}5}$
    & $\mathrm{F}_{\mathrm{1,mic}\text{-}14}$
    & $\mathrm{F}_{\mathrm{1,mac}\text{-}14}$
    & $\mathrm{F}_{1}\text{-}\mathrm{SRR}\text{-}\mathrm{BERT}$
    & $\mathrm{RG}\text{-}\mathrm{F}_{1}$ \\
    \midrule
    \rowcolor{TableRowBlue}
    & SFT & \metricci{0.576\sigmark{**}}{0.554}{0.600} & \metricci{0.506\sigmark{**}}{0.483}{0.527} & \metricci{0.537\sigmark{**}}{0.521}{0.555} & \metricci{0.412\sigmark{**}}{0.393}{0.428} & \metricci{0.503\sigmark{*}}{0.495}{0.517} & \metricci{0.243\sigmark{**}}{0.235}{0.251} \\
    \cmidrule[0.50pt]{2-8}
    \rowcolor{TableRowGrey}
    & SFT w/ CoT & \metricci{0.590\sigmark{**}}{0.569}{0.613} & \metricci{0.519\sigmark{**}}{0.501}{0.538} & \metricci{0.545\sigmark{**}}{0.530}{0.564} & \metricci{0.412\sigmark{**}}{0.383}{0.423} & \metricci{0.487\sigmark{**}}{0.474}{0.499} & \metricci{0.227\sigmark{**}}{0.219}{0.235} \\
    \cmidrule[0.50pt]{2-8}
    \rowcolor{TableRowBlue}
    & SFT w/ MCoT & \metricci{0.601\sigmark{**}}{0.576}{0.616} & \metricci{0.523\sigmark{**}}{0.506}{0.545} & \metricci{0.550\sigmark{**}}{0.537}{0.568} & \metricci{0.410\sigmark{**}}{0.391}{0.425} & \metricci{0.501\sigmark{**}}{0.488}{0.510} & \metricci{0.236\sigmark{**}}{0.228}{0.244} \\
    \cmidrule[0.50pt]{2-8}
    \rowcolor{TableARIVE}
    \multirow{-4}{*}{MIMIC-CXR} & ARIVE & \metricci{\textbf{0.652}}{0.630}{0.672} & \metricci{\textbf{0.555}}{0.536}{0.571} & \metricci{\textbf{0.603}}{0.586}{0.617} & \metricci{\textbf{0.437}}{0.416}{0.453} & \metricci{\textbf{0.515}}{0.503}{0.526} & \metricci{\textbf{0.271}}{0.263}{0.279} \\
    \midrule
    \rowcolor{TableRowBlue}
    & SFT & \metricci{0.546\sigmark{*}}{0.490}{0.599} & \metricci{0.482\sigmark{*}}{0.427}{0.530} & \metricci{0.497\sigmark{**}}{0.462}{0.531} & \metricci{0.366\sigmark{*}}{0.332}{0.394} & \metricci{0.408\sigmark{**}}{0.375}{0.443} & \metricci{0.193}{0.176}{0.211} \\
    \cmidrule[0.50pt]{2-8}
    \rowcolor{TableRowGrey}
    & SFT w/ CoT & \metricci{0.545\sigmark{**}}{0.483}{0.583} & \metricci{0.497\sigmark{*}}{0.443}{0.536} & \metricci{0.502\sigmark{**}}{0.471}{0.531} & \metricci{0.374\sigmark{*}}{0.339}{0.396} & \metricci{0.394\sigmark{**}}{0.368}{0.429} & \metricci{0.183\sigmark{**}}{0.165}{0.199} \\
    \cmidrule[0.50pt]{2-8}
    \rowcolor{TableRowBlue}
    & SFT w/ MCoT & \metricci{0.555}{0.519}{0.621} & \metricci{0.512}{0.474}{0.591} & \metricci{0.514\sigmark{**}}{0.470}{0.540} & \metricci{0.377\sigmark{*}}{0.337}{0.407} & \metricci{0.410\sigmark{**}}{0.369}{0.434} & \metricci{0.188\sigmark{*}}{0.173}{0.203} \\
    \cmidrule[0.50pt]{2-8}
    \rowcolor{TableARIVE}
    \multirow{-4}{*}{MIMIC-ABN} & ARIVE & \metricci{\textbf{0.581}}{0.529}{0.631} & \metricci{\textbf{0.541}}{0.485}{0.590} & \metricci{\textbf{0.545}}{0.514}{0.574} & \metricci{\textbf{0.408}}{0.372}{0.433} & \metricci{\textbf{0.441}}{0.405}{0.477} & \metricci{\textbf{0.207}}{0.192}{0.222} \\
    \bottomrule
  \end{tabular}
  \end{adjustbox}
  \endgroup
\end{table*}

\subsubsection{Contributions of ACCO and its components}
\label{sec:reward_component_ablation}
With MCoT supervision fixed, adding ACCO improved all six tabulated clinical-efficacy measures on both datasets (Table~\ref{tab:training_ablation}). Relative to SFT w/ MCoT, the primary 14-observation macro $\mathrm{F}_{1}$ increased by 6.6\% on MIMIC-CXR (0.437 versus 0.410) and by 8.2\% on MIMIC-ABN (0.408 versus 0.377), and GREEN increased from 0.320 to 0.360 on MIMIC-CXR (Fig.~\ref{fig:cxr_evaluation_composite}f). All MIMIC-CXR comparisons were significant at $p < 0.01$, and four of the six tabulated MIMIC-ABN comparisons were significant at $p < 0.05$. These gains spanned finding accuracy, fine-grained clinical content and entity--relation structure. Combined with MCoT supervision, ACCO yielded a net gain over report-only SFT in CheXbert $\mathrm{F}_{1}$ for nine of the 14 observations on MIMIC-CXR, with equal performance for Pleural Other and the largest gains for Edema and Atelectasis (Fig.~\ref{fig:cxr_evaluation_composite}d).

To assess the contribution of each ACCO component, we removed each reward, or the inverse-frequency weighting, in turn and evaluated report accuracy, anatomical grounding, reasoning-stage clinical content and reasoning--report consistency on MIMIC-ABN and its Gold subset (Table~\ref{tab:beta_ablation}). Removing the anatomical-grounding reward $R_{\mathrm{BBOX}}$ reduced every reported measure, supporting its contribution to both anatomical grounding and downstream clinical content. Removing the reasoning-stage clinical reward $R^{\mathrm{rea}}_{\mathrm{CXB}}$ reduced reasoning-stage and final-report $\mathrm{F}_{1}$ on the Gold subset and lowered reasoning--report consistency macro $\mathrm{F}_{1}$ from 0.887 to 0.782. These concurrent declines support the role of intermediate supervision in preserving clinical content in the final report.

Removing the report-stage clinical reward $R^{\mathrm{ans}}_{\mathrm{CXB}}$ produced the largest reduction in final-report micro $\mathrm{F}_{1}$ among the ACCO component ablations, both on the Gold subset (0.463 versus 0.505, an 8.3\% relative decline) and on the full MIMIC-ABN test set (0.526 versus 0.545). However, region-selection $\mathrm{F}_{1}$ and mIoU increased without this reward, with mIoU reaching 0.493 compared with 0.474 for full ARIVE. This divergence indicates that better anatomical grounding alone does not ensure more accurate findings in the final report. Removing the RadGraph reward $R_{\mathrm{RG}}$ lowered both RadGraph $\mathrm{F}_{1}$ (0.200 versus 0.207) and $\mathrm{F}_{1}\text{-}\mathrm{SRR}\text{-}\mathrm{BERT}$ (0.417 versus 0.441) on MIMIC-ABN, supporting its contribution to preserving clinical entities, their relations and fine-grained findings.

Removing inverse-frequency weighting slightly improved overall anatomical grounding but reduced macro-averaged $\mathrm{F}_{1}$ for both reasoning-stage and final-report content on the Gold subset, with the latter falling from 0.444 to 0.400. Final-report micro $\mathrm{F}_{1}$ was unchanged (0.505), suggesting that the losses were concentrated in less frequent observations, which the weighting is designed to emphasize. Likewise, the gain in overall grounding may reflect greater emphasis on frequently annotated regions once weighting is removed. Together, these ablations show that anatomical grounding and clinical efficacy do not necessarily improve in parallel, underscoring the need to optimize them jointly.

\begin{table*}[!ht]
  \centering
  \caption{MCoT supervision and ACCO component ablations on MIMIC-ABN and its 142-example Chest ImaGenome Gold subset. CheXbert $\mathrm{F}_{\mathrm{1,14}}$ values are micro/macro averages across 14 observations; superscripts $\mathrm{rea}$, $\mathrm{ans}$ and $\mathrm{con}$ denote reasoning-stage content, final-report content and reasoning--report consistency, respectively. Bold marks the highest value for each metric across all settings; micro/macro maxima are identified separately.}
  \label{tab:beta_ablation}
  \begingroup
  \renewcommand{\arraystretch}{1.20}
  \setlength{\tabcolsep}{4.5pt}
  \resizebox{\textwidth}{!}{%
  \begin{tabular}{lcccccccc}
    \toprule
    \rowcolor{TableHeader}
    & \multicolumn{3}{c}{\textbf{MIMIC-ABN test set}}
    & \multicolumn{5}{c}{\textbf{Chest ImaGenome Gold subset}} \\
    \cmidrule(lr){2-4}\cmidrule(lr){5-9}
    \rowcolor{TableHeader}
    \multirow{-2}{*}{\textbf{Training setting}}
    & \begin{tabular}[c]{@{}c@{}}$\mathrm{F}_{\mathrm{1,14}}^{\,\mathrm{ans}}$\\(mic/mac)\end{tabular}
    & $\mathrm{F}_{1}\text{-}\mathrm{SRR}\text{-}\mathrm{BERT}$
    & $\mathrm{RG}\text{-}\mathrm{F}_{1}$
    & \begin{tabular}[c]{@{}c@{}}Region-\\selection $\mathrm{F}_{1}$\end{tabular}
    & $\mathrm{mIoU}$
    & \begin{tabular}[c]{@{}c@{}}$\mathrm{F}_{\mathrm{1,14}}^{\,\mathrm{rea}}$\\(mic/mac)\end{tabular}
    & \begin{tabular}[c]{@{}c@{}}$\mathrm{F}_{\mathrm{1,14}}^{\,\mathrm{ans}}$\\(mic/mac)\end{tabular}
    & \begin{tabular}[c]{@{}c@{}}$\mathrm{F}_{\mathrm{1,14}}^{\,\mathrm{con}}$\\(mic/mac)\end{tabular} \\
    \midrule
    \rowcolor{TableSection}
    \multicolumn{9}{c}{\textbf{\textit{Supervised fine-tuning comparison}}} \\
    \midrule
    \rowcolor{TableRowBlue}
    SFT w/ CoT  & 0.502/0.374 & 0.394 & 0.183 & 0.579 & 0.430 & 0.369/0.310 & 0.455/0.374 & 0.714/0.728 \\
    \midrule[0.50pt]
    \rowcolor{TableRowGrey}
    SFT w/ MCoT  & 0.514/0.377 & 0.410 & 0.188 & 0.605 & 0.474 & 0.463/0.413 & 0.472/0.411 & 0.902/0.894 \\
    \midrule
    \rowcolor{TableSection}
    \multicolumn{9}{c}{\textbf{\textit{ACCO ablations}}} \\
    \midrule
    \rowcolor{TableARIVE}
    ARIVE  & \metricbold{0.545}/\metricbold{0.408} & \metricbold{0.441} & 0.207 & 0.603 & 0.474 & \metricbold{0.513}/\metricbold{0.451} & \metricbold{0.505}/\metricbold{0.444} & \metricbold{0.945}/0.887 \\
    \midrule[0.50pt]
    \rowcolor{TableRowBlue}
    w/o $R_{\mathrm{BBOX}}$  & 0.539/0.396 & 0.421 & 0.196 & 0.591 & 0.455 & 0.501/0.421 & 0.492/0.400 & 0.914/0.856 \\
    \midrule[0.50pt]
    \rowcolor{TableRowGrey}
    w/o $R^{\mathrm{rea}}_{\mathrm{CXB}}$  & 0.539/0.400 & \metricbold{0.441} & 0.211 & 0.612 & 0.479 & 0.484/0.381 & 0.481/0.389 & 0.894/0.782 \\
    \midrule[0.50pt]
    \rowcolor{TableRowBlue}
    w/o $R^{\mathrm{ans}}_{\mathrm{CXB}}$  & 0.526/0.388 & 0.428 & \metricbold{0.212} & \metricbold{0.618} & \metricbold{0.493} & 0.500/0.426 & 0.463/0.392 & 0.929/\metricbold{0.911} \\
    \midrule[0.50pt]
    \rowcolor{TableRowGrey}
    w/o $R_{\mathrm{RG}}$  & 0.542/0.388 & 0.417 & 0.200 & 0.608 & 0.477 & 0.502/0.390 & 0.485/0.379 & 0.922/0.730 \\
    \midrule[0.50pt]
    \rowcolor{TableRowBlue}
    w/o weighting  & 0.538/0.400 & 0.428 & 0.208 & 0.608 & 0.481 & 0.496/0.379 & \metricbold{0.505}/0.400 & 0.940/0.884 \\
    \bottomrule
  \end{tabular}}
  \endgroup
\end{table*}

\label{sec:localization_analysis}
The 636-example MIMIC-CXR Gold subset provided complementary evidence on anatomical grounding (Fig.~\ref{fig:cxr_evaluation_composite}g). On this subset, region-selection $\mathrm{F}_{1}$ and mIoU improved progressively from text-only CoT to MCoT and then to full ARIVE, whereas on the smaller MIMIC-ABN Gold subset, overall grounding was similar before and after ACCO (Table~\ref{tab:beta_ablation}). The frequently annotated right and left lungs showed similarly high region-selection performance in all three settings. The less frequently annotated right upper lung zone, by contrast, differed markedly: its region-selection $\mathrm{F}_{1}$ reached 0.226 with ARIVE, compared with 0.083 with MCoT and 0.043 with text-only CoT. These results suggest that ACCO improves region selection for underrepresented anatomy.

Taken together, the staged comparisons and component ablations reveal complementary roles for MCoT and ACCO. Relative to text-only CoT, MCoT improved the measured clinical content of intermediate reasoning and its consistency with final reports by incorporating regional visual evidence. Adding ACCO improved report-level clinical-content metrics and entity–relation agreement, with gains in macro-averaged observation scores and in region selection for some underrepresented anatomy. We next assessed generated reports and MCoT trajectories through radiologist grading and examined whether access to MCoT supported report review.

\subsection{Clinical evaluation of generated reports and MCoT trajectories}
\label{sec:clinician_evaluation}

\subsubsection{Blinded grading of report quality}
\label{sec:blinded_report_assessment}

\begin{figure}[!htb]
\centering
\includegraphics[width=\textwidth]{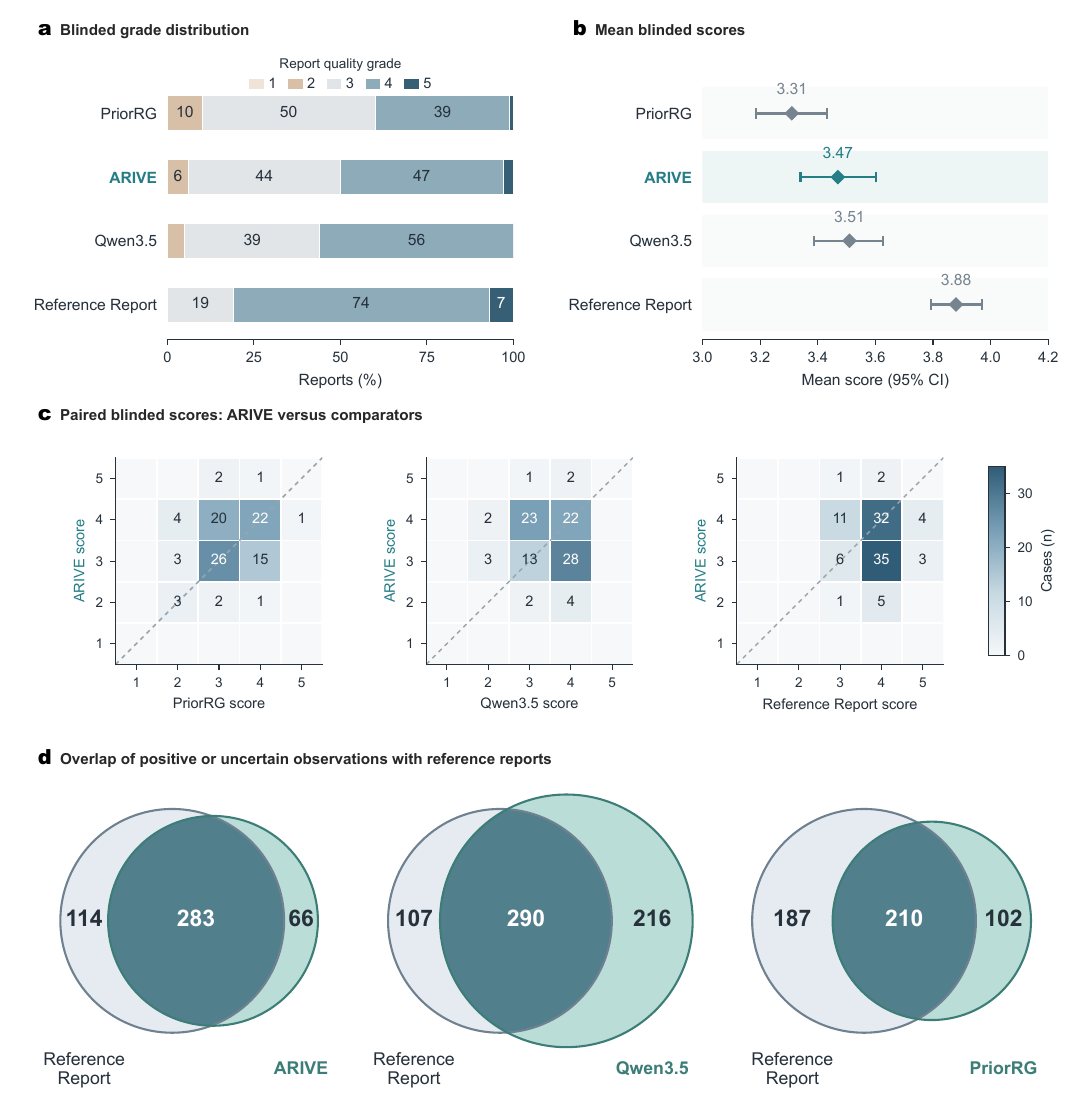}
\caption{Report-quality grading and observation overlap with reference reports. (a), Blinded grade distributions assigned by the radiologist for 100 randomly selected cases from 71 patients. Scores range from 1 (unusable) to 5 (directly usable). (b), Mean report-quality scores with 95\% confidence intervals. (c), Paired scores for ARIVE versus PriorRG, Qwen3.5 and reference reports, from left to right. Cells show case counts and dashed diagonals indicate equal scores. (d), Observation overlap between generated and reference reports. Left, ARIVE; center, Qwen3.5; right, PriorRG. Numbers indicate observations shared by the generated and reference reports or present in only one.}
\label{fig:blinded_report_quality}
\end{figure}

Following previous expert evaluations~\cite{chen2025deep}, an experienced radiologist graded reference reports and reports generated by ARIVE, PriorRG and Qwen3.5 for 100 randomly selected MIMIC-CXR cases from 71 patients (Fig.~\ref{fig:blinded_report_quality}a--c). These two comparators were the strongest on complementary measures: PriorRG led the comparators on RadGraph $\mathrm{F}_{1}$, $\mathrm{F}_{1}\text{-}\mathrm{SRR}\text{-}\mathrm{BERT}$ and GREEN, and Qwen3.5 on the primary 14-observation macro $\mathrm{F}_{1}$ (Table~\ref{tab:main_results}). Reports were presented with their sources concealed, together with the current radiograph, the clinical indication and, when available, the prior image and report. Each report was graded for clinical accuracy on a five-point scale from 1 (unusable) to 5 (directly usable).

ARIVE reports more often required at most minor edits (grades 4--5) than PriorRG reports (50\% versus 40\%), although less often than Qwen3.5 reports (56\%). The corresponding proportion for reference reports was 81\% (Fig.~\ref{fig:blinded_report_quality}a). ARIVE's mean score was higher than that of PriorRG (3.47 versus 3.31) and similar to that of Qwen3.5 (3.51; Fig.~\ref{fig:blinded_report_quality}b). Reference reports received the highest mean score (3.88), exceeding those of all three models. In paired comparisons, ARIVE received a higher grade than PriorRG in 30 cases, a lower grade in 19 and an equal grade in 51. Against Qwen3.5, the corresponding counts were 31, 34 and 35 (Fig.~\ref{fig:blinded_report_quality}c).

To examine report content alongside these grades, we used a large language model (DeepSeek-V4.1-Flash~\cite{deepseekai2026deepseekv41flash}) to extract observation states from the same 400 reports and compared each generated report with its reference (Fig.~\ref{fig:blinded_report_quality}d). ARIVE's higher mean score than PriorRG was accompanied by greater coverage of reference observations and fewer observations absent from the references. ARIVE and Qwen3.5, despite their similar mean scores, differed in what they reported. ARIVE recovered slightly fewer reference observations than Qwen3.5 (71.3\% versus 73.0\%), but 81.1\% of its reported observations were supported by the references, compared with 57.3\% for Qwen3.5. Qwen3.5 reported more broadly, capturing slightly more reference findings but also more observations without reference support (216 versus 66), whereas ARIVE's reports were more selective, trading slightly lower coverage for higher agreement with the references.

\subsubsection{Radiologist grading of MCoT trajectories}
\label{sec:mcot_assisted_clinician_review}

To assess the MCoT trajectories generated before ARIVE's reports, the radiologist graded trajectories for 100 randomly selected cases from 80 patients, sampled separately from the cases used for report grading. Each trajectory was graded from 1 (lowest) to 5 (highest) in four domains: clinical factual correctness, anatomical localization accuracy, diagnostic traceability (whether the main judgments in the final report could be traced to the trajectory) and expression quality, yielding 400 grades in total.

\begin{figure}[!p]
\centering
\includegraphics[width=\textwidth,height=0.78\textheight,keepaspectratio]{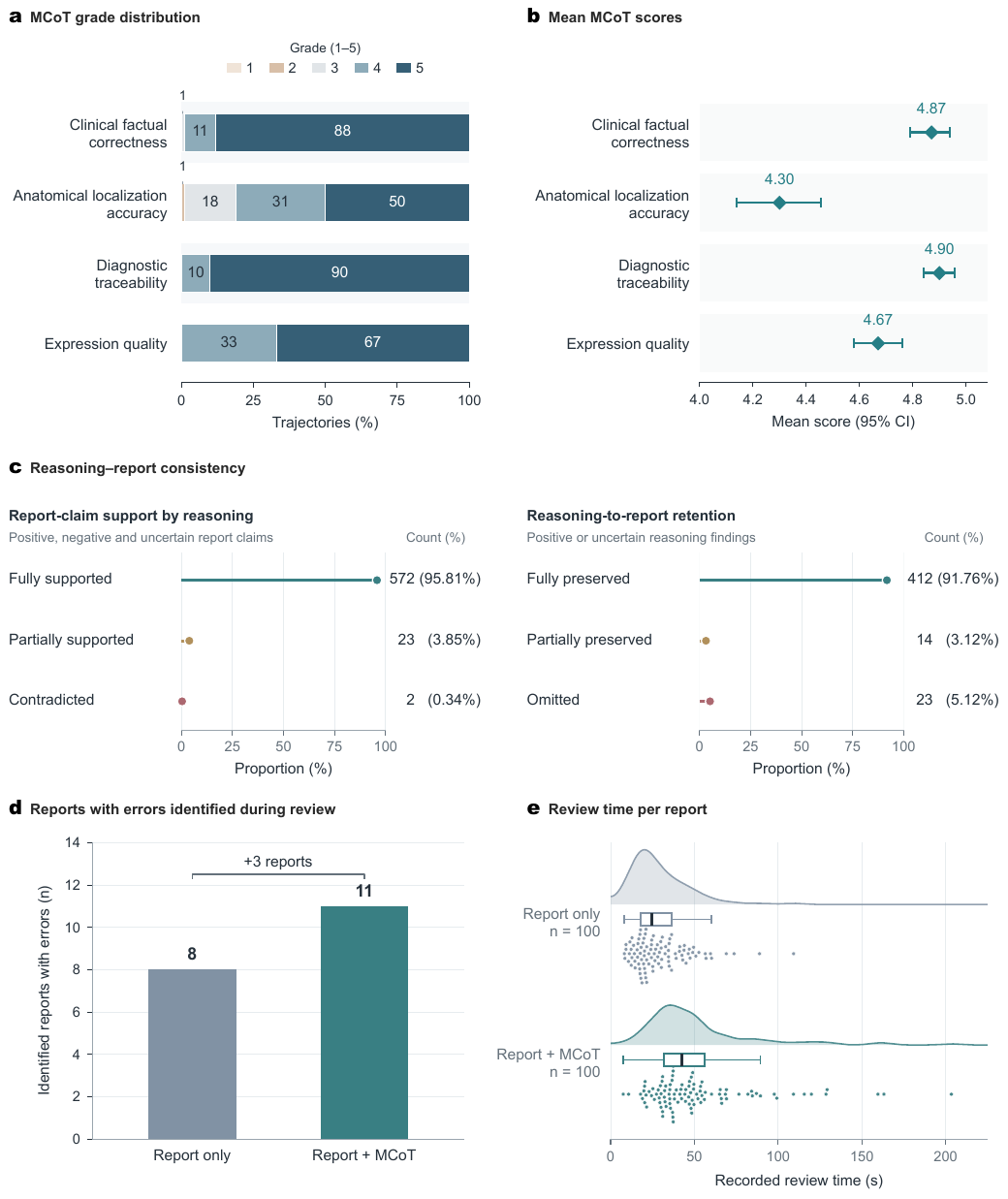}
\caption{MCoT trajectory quality, reasoning--report consistency and report review. (a), Grade distributions assigned by the radiologist across four domains for 100 randomly selected cases from 80 patients. Scores range from 1 (lowest) to 5 (highest). (b), Mean MCoT scores with 95\% confidence intervals. (c), Bidirectional reasoning--report consistency. Left, support from MCoT trajectories for report claims ($n = 597$; positive, negative or uncertain). Right, retention of MCoT findings in the final reports ($n = 449$; positive or uncertain). (d), Number of reports judged to contain errors during paired review without and with MCoT (8 and 11 of 100 reports, respectively). (e), Review time per report under each condition. Points show individual review times and shaded curves show kernel density estimates. Boxes span the interquartile range (IQR), with medians marked. Whiskers extend to the most extreme observations within 1.5 IQR of the box edges.}
\label{fig:mcot_reasoning_scores}
\end{figure}

Ratings were consistently high in three domains: clinical factual correctness, diagnostic traceability and expression quality received grades of 4 or 5 in 99\%, 100\% and 100\% of trajectories, respectively (Fig.~\ref{fig:mcot_reasoning_scores}a). Anatomical localization accuracy was more variable, with 81\% of trajectories graded 4 or 5, 18\% graded 3 and 1\% graded 2. The top grade was assigned for diagnostic traceability in 90\% of cases but for anatomical localization accuracy in only 50\%. Mean scores followed the same pattern: diagnostic traceability (4.90; 95\% CI, 4.84--4.96) and clinical factual correctness (4.87; 4.79--4.94) scored highest, followed by expression quality (4.67; 4.58--4.76) and anatomical localization accuracy (4.30; 4.14--4.46; Fig.~\ref{fig:mcot_reasoning_scores}b). The radiologist attributed the lower localization scores partly to incomplete coverage of the left lung and heart in the anatomical crops.

To complement the traceability grades, we quantified reasoning--report consistency at the level of individual claims and findings (Fig.~\ref{fig:mcot_reasoning_scores}c). Using the same large language model, we extracted claims from the final reports and findings from the MCoT trajectories, and then measured the proportion of report claims supported by the reasoning and the proportion of reasoning findings retained in the reports. Partial support denoted agreement on the core finding but not on all of its attributes. Of 597 positive, negative or uncertain report claims, 572 (95.8\%) were fully supported by the corresponding trajectories, 23 (3.9\%) were partially supported (14 with differences in temporal attributes, 7 in certainty, 1 in location and 1 in morphology) and 2 (0.3\%) were contradicted (one pleural effusion claim and one pulmonary vascular congestion claim). Nearly all report claims could therefore be traced to the trajectories, consistent with the high traceability grades.

In the reverse direction, partial retention denoted that a report preserved the core finding but lost or altered clinically relevant information. Of 449 positive or uncertain reasoning findings, 412 (91.8\%) were fully retained in the reports, 14 (3.1\%) were partially retained (temporal information changed or missing in 9 and certainty changed in 5) and 23 (5.1\%) were omitted (8 involving devices, 5 pneumonia, 4 pulmonary edema, 3 consolidation, 2 atelectasis and 1 pleural effusion). Most positive or uncertain findings in the trajectories therefore carried through to the final reports, indicating strong reasoning--report consistency in both directions.

\subsubsection{MCoT-assisted review of generated reports}
\label{sec:mcot_assisted_report_review}

To assess whether access to MCoT could help reviewers identify errors in generated reports, the radiologist reviewed the ARIVE reports for the same 100 cases in two sessions. In the first session, each report was reviewed alongside its source radiograph. After a two-week interval intended to reduce recall, the same reports and radiographs were reviewed with the corresponding MCoT trajectories available. In each session, we recorded which reports were judged to contain clinically significant errors (false or omitted findings) and the review time for each report.

The radiologist identified errors in 8 reports without MCoT and in 11 with MCoT (Fig.~\ref{fig:mcot_reasoning_scores}d). Errors were identified again in all 8 reports flagged in the first session, and the MCoT-assisted session identified 3 additional reports containing errors. Median review time increased from 24.5~s (interquartile range (IQR), 18.0--36.4~s) without MCoT to 42.5~s (IQR, 31.8--56.1~s) with MCoT (Fig.~\ref{fig:mcot_reasoning_scores}e). These paired results provide preliminary evidence that access to MCoT may support error identification, although review time was longer. The case studies below illustrate how regional evidence shapes individual trajectories and how findings and errors can be traced into the final reports.

\subsection{Case studies of generated reports and diagnostic trajectories}
\label{sec:qualitative_analysis}

\subsubsection{Comparison of generated reports}
\label{sec:report_quality_cases}

\begin{figure*}[!p]
\centering
\includegraphics[width=\textwidth,height=0.95\textheight,keepaspectratio]{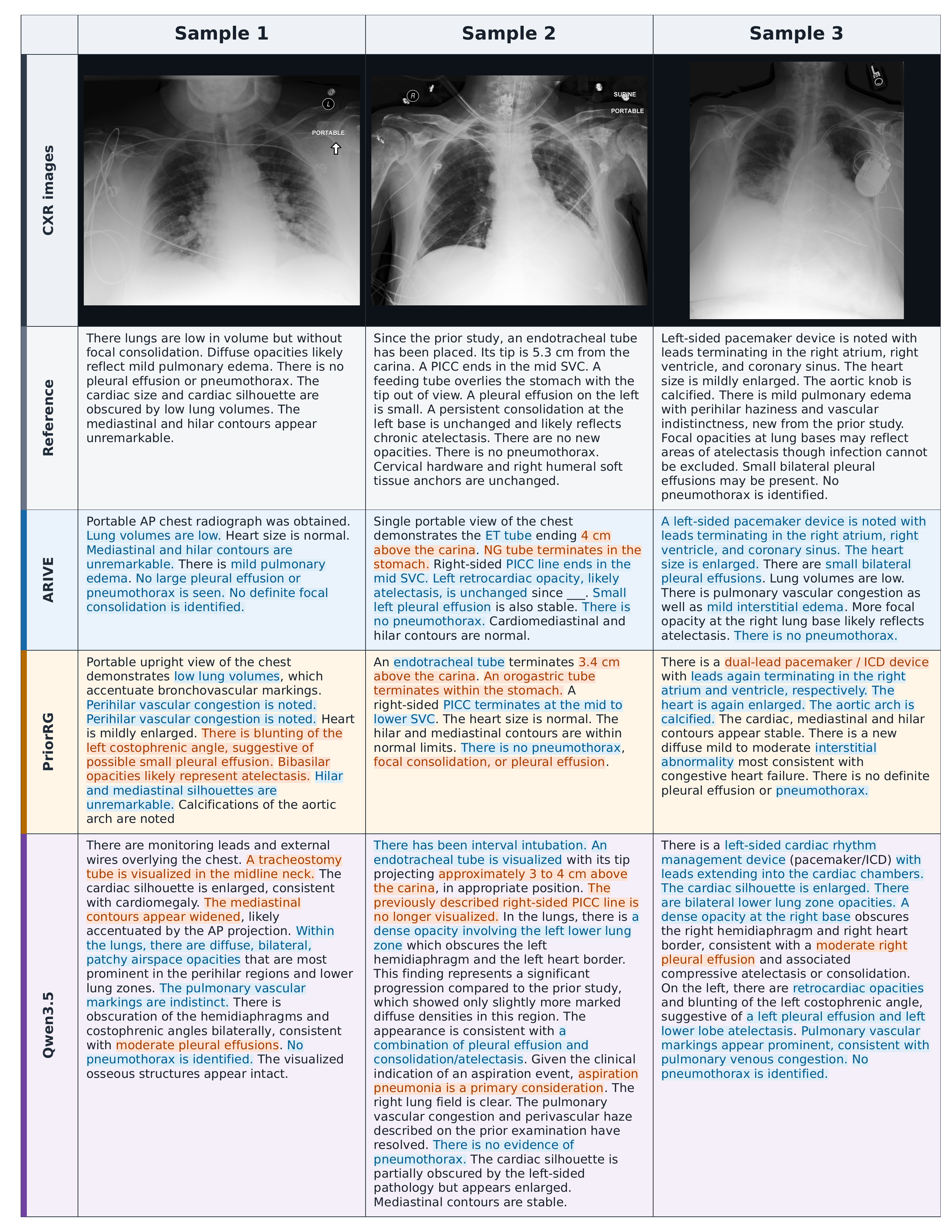}
\caption{Final-report comparison in three selected MIMIC-CXR cases. Each column presents the radiograph, reference report and reports generated by ARIVE, PriorRG and Qwen3.5. Blue and vermilion highlights indicate concordant and unsupported findings, respectively. In these examples, ARIVE captures the principal findings and key negative observations while avoiding several unsupported or overstated findings produced by the comparators.}
\label{fig:report_comparison}
\end{figure*}

We compared reports generated by ARIVE, PriorRG and Qwen3.5 with the reference reports in three selected cases (Fig.~\ref{fig:report_comparison}). In Sample~1, ARIVE correctly reported the low lung volumes and mild pulmonary edema, together with the absence of focal consolidation, pneumothorax and a large pleural effusion. PriorRG instead added mild cardiomegaly, bibasal atelectasis and a possible left pleural effusion, whereas Qwen3.5 introduced a tracheostomy tube and moderate bilateral pleural effusions. In Sample~2, ARIVE retained the endotracheal tube, peripherally inserted central catheter (PICC), unchanged retrocardiac atelectasis, small left pleural effusion and absence of pneumothorax. PriorRG omitted both the pleural effusion and the retrocardiac abnormality, while Qwen3.5 incorrectly reported removal of the PICC and interpreted the chronic opacity as progressive aspiration pneumonia. ARIVE, however, erroneously placed the feeding-tube tip in the stomach, whereas the reference stated that the tip was outside the field of view. In Sample~3, ARIVE preserved the three-lead pacemaker configuration, cardiac enlargement, pulmonary edema, small bilateral pleural effusions, basilar atelectasis and absence of pneumothorax. PriorRG described only two leads and omitted the pleural effusions and basilar opacity, whereas Qwen3.5 overstated the right pleural effusion as moderate. ARIVE omitted the aortic calcification and the differential considerations for the left basilar opacity but avoided these comparator errors.

Across these cases, ARIVE captured the major abnormalities, devices and key negative findings with fewer unsupported details than the comparators, while still making occasional errors and omissions. These examples complement the quantitative results by illustrating both concordant and unsupported content within individual reports.

\begin{figure*}[!p]
\centering
\includegraphics[width=0.98\textwidth,height=0.95\textheight,keepaspectratio]{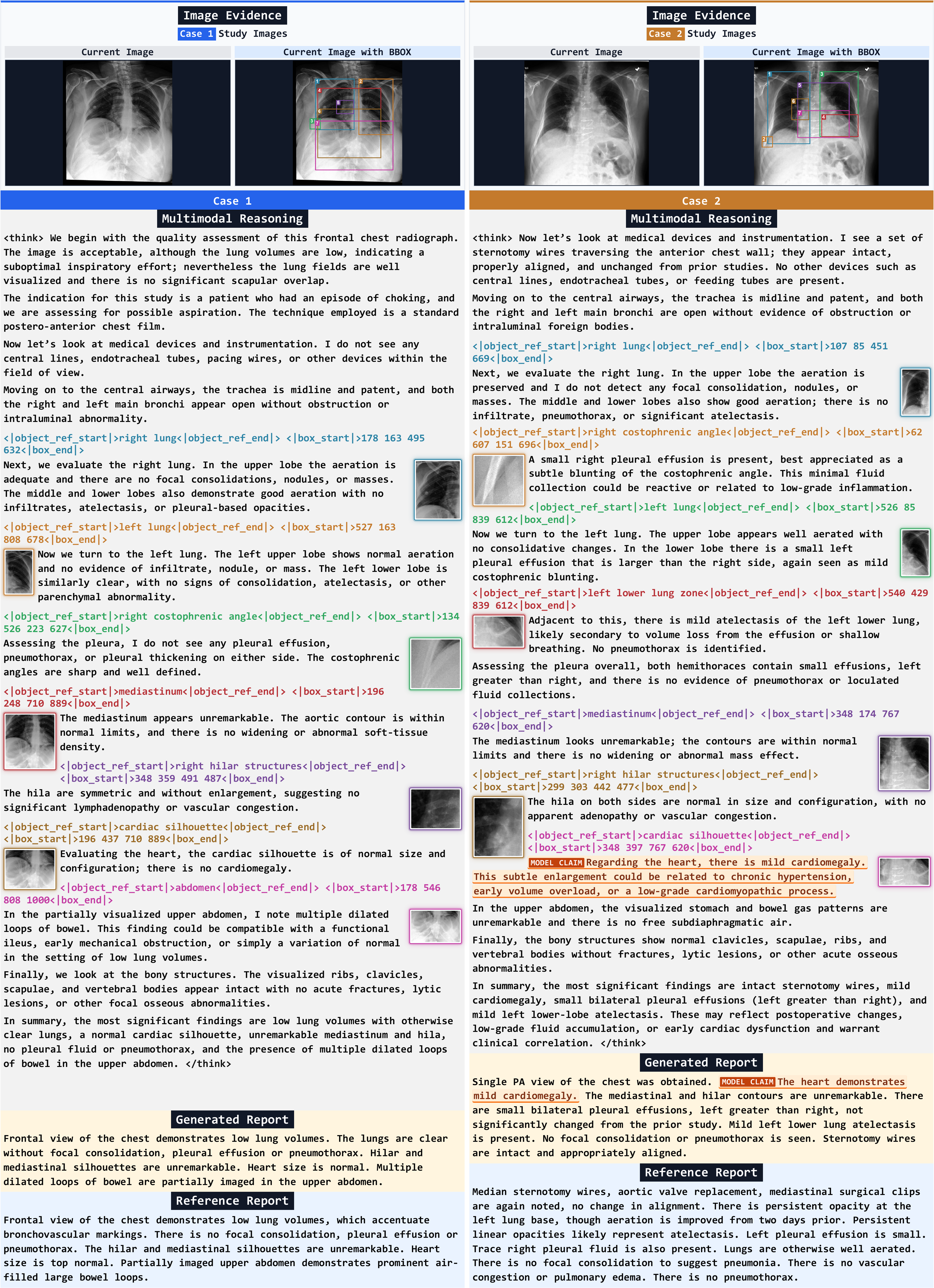}
\caption{Inspectable ARIVE trajectories in two selected MIMIC-CXR cases. Each case pairs the input radiograph and predicted anatomical boxes with the requested regions, MCoT trajectory, generated report and reference report. Case~1 shows a trajectory broadly concordant with the reference report. Case~2 traces an unsupported claim of mild cardiomegaly from intermediate reasoning into the final report.}
\label{fig:mcot_quality_analysis}
\end{figure*}

\subsubsection{Tracing findings and errors through MCoT trajectories}
\label{sec:trajectory_case_studies}

Report-level fidelity alone does not reveal how individual findings arise during the diagnostic process. We therefore examined two further cases (Fig.~\ref{fig:mcot_quality_analysis}), each pairing regional image evidence with the corresponding intermediate interpretations, final report and reference report.

Case~1 illustrates a clinically concordant trajectory. ARIVE inspected the lung, pleural, cardiac and upper-abdominal regions and identified low lung volumes, clear lungs, normal heart size and dilated bowel loops; these interpretations were carried into the final report and closely matched the reference findings. Case~2 shows how reasoning--report consistency makes the origin of an unsupported finding traceable. The generated report described mild cardiomegaly, which was absent from the reference report. This claim first appeared in an earlier \textsc{Model Claim} step that characterized the heart as mildly enlarged, and the trajectory displayed the cardiac-silhouette region associated with it. A reviewer can therefore check the intermediate claim against the image evidence and follow its propagation into the final report.

Together, these cases illustrate how multimodal trajectories make reported findings inspectable by linking regional visual evidence, intermediate interpretations and final report statements. The same linkage allows an unsupported claim to be traced from its first appearance in the reasoning to its inclusion in the final report.

\section{Discussion}
\label{sec:discussion}
In this study, we developed ARIVE, an RRG framework that interleaves anatomical inspection with diagnostic interpretation before composing the final report. ARIVE achieved the highest scores across the clinical-efficacy measures on both MIMIC-CXR and the abnormality-centered MIMIC-ABN, with statistically significant differences in most comparisons (Table~\ref{tab:main_results} and Fig.~\ref{fig:cxr_evaluation_composite}c). Evaluation on the Chest ImaGenome Gold subset further supported its ability to select and localize anatomical regions during reasoning (Fig.~\ref{fig:cxr_evaluation_composite}g). In blinded grading, ARIVE's mean report-quality score was numerically higher than PriorRG's and similar to Qwen3.5's (Fig.~\ref{fig:blinded_report_quality}), and the radiologist rated its MCoT trajectories highly for diagnostic traceability. At the claim level, 95.8\% of report claims were fully supported by the corresponding trajectories (Fig.~\ref{fig:mcot_reasoning_scores}a--c). With MCoT available, the radiologist identified errors in more reports (11 versus 8 of 100; Fig.~\ref{fig:mcot_reasoning_scores}d), providing preliminary evidence that trajectories can support report review. Together, these findings indicate that inspectable diagnostic reasoning can be built into report generation without sacrificing clinical efficacy metrics.

Among the baselines, the general-purpose and generalist medical MLLMs were strongest on observation-level accuracy as measured by CheXbert, whereas the task-specific RRG models tended to achieve greater reference agreement in clinical entities and their relations; ARIVE led on both types of measure. Its advantage over Qwen3.5 was largest on the CheXbert measures dominated by common observations and smallest on the primary 14-observation macro $\mathrm{F}_{1}$. ARIVE achieved these results with 4.85 billion parameters, far fewer than Qwen3.5 (397 billion total and 17 billion active parameters) and also used fewer parameters than MAIRA-2 and Libra. Generating both an MCoT trajectory and a final report took a mean of 1.13~s per sample on a single NVIDIA A100 GPU, comparable to the report-only inference times of MedGemma-1.5 and MAIRA-2 but longer than those of the smaller MLRG and PriorRG models (Supplementary Table~\ref{supp-tab:mimic_inference_profile}). ARIVE thus combines strong clinical efficacy with inspectable reasoning at a moderate model scale and inference cost.

The staged ablations help explain these gains. Text-only CoT made intermediate diagnostic statements explicit, but its gains on some CheXbert measures came with lower $\mathrm{F}_{1}\text{-}\mathrm{SRR}\text{-}\mathrm{BERT}$, RadGraph $\mathrm{F}_{1}$ and GREEN on MIMIC-CXR (Table~\ref{tab:training_ablation} and Fig.~\ref{fig:cxr_evaluation_composite}e,f). Thus, explicit intermediate statements alone did not consistently improve report performance. Interleaving anatomical inspection through MCoT improved most clinical-efficacy measures relative to text-only CoT, together with reasoning-stage observation accuracy and reasoning--report consistency (Tables~\ref{tab:training_ablation} and~\ref{tab:beta_ablation}). One possible explanation consistent with these results is that later steps of long textual reasoning rely increasingly on preceding text rather than on the image~\cite{sun2025mitigating,cheng2026visual}, whereas regional inspection renews visual support as the trajectory evolves and thereby limits the propagation of unsupported assumptions~\cite{jin2026look}. Prior work has linked report findings to localized image evidence for grounding, verification or staged interpretation~\cite{bannur2024maira,wu2026universal,jin2026grounded,zhang2026generalised,hashmi2026anatomix,chen2026interpretable}. ARIVE incorporates regional evidence iteratively into the generation sequence, where it conditions subsequent interpretation before report composition. MCoT alone, however, remained below report-only SFT on RadGraph $\mathrm{F}_{1}$ and GREEN and these scores exceeded those of report-only SFT only after ACCO was added. 

ACCO addressed a complementary limitation: the token-level SFT objective does not directly reward accurate grounding or clinically correct content, nor does it emphasize underrepresented observations and anatomical regions. Adding ACCO to MCoT-based training increased scores on every clinical-efficacy measure on both datasets (Table~\ref{tab:training_ablation} and Fig.~\ref{fig:cxr_evaluation_composite}f), and its component ablations supported distinct contributions from the rewards (Table~\ref{tab:beta_ablation}). Removing the anatomical-grounding reward impaired all evaluated outcomes, whereas removing the report-stage clinical reward $R^{\mathrm{ans}}_{\mathrm{CXB}}$ improved grounding yet caused the largest decline in final-report micro $\mathrm{F}_{1}$ among the reward ablations. This divergence suggests that localization quality alone is an insufficient proxy for report accuracy. Inverse-frequency weighting involved a related trade-off, accepting a slight loss in overall grounding for better macro-averaged reasoning and report performance. These results support jointly optimizing anatomical grounding, reasoning--report consistency and final-report clinical fidelity, with frequency weighting to address imbalances in clinical observations and anatomical regions.

The radiologist's grades add a complementary perspective. Although ARIVE exceeded Qwen3.5 on every automated clinical-efficacy measure, the two received similar mean grades (3.47 versus 3.51). One plausible reason is that reference-based metrics penalize observations absent from the reference, which Qwen3.5 reported more than three times as often as ARIVE (216 versus 66; Fig.~\ref{fig:blinded_report_quality}d), while a radiologist reviewing the image may accept some of them, particularly because the reference reports were themselves imperfect (mean grade, 3.88). Automated metrics and expert grading therefore capture different aspects of report quality.

For report review, the trajectories make intermediate interpretations and their associated regional evidence available for inspection. With MCoT available, the radiologist identified three additional reports containing errors, while review time increased (Fig.~\ref{fig:mcot_reasoning_scores}d,e). This finding suggests a potential trade-off between error identification and review time. Inspectability, however, does not guarantee correctness. As Case~2 illustrates, an unsupported finding can propagate coherently from the trajectory into the report (Fig.~\ref{fig:mcot_quality_analysis}), and a fluent rationale could lend such an incorrect finding unwarranted credibility. MCoT trajectories should therefore be treated as generated interpretations to be checked against image evidence, and ARIVE is intended to support rather than replace clinical judgment.

Several limitations define the scope of these findings. First, anatomical supervision was sparse for some regions: the upper and mid lung zones and apical regions together accounted for only approximately 2.8\% of all bounding-box annotations (Supplementary Table~\ref{supp-tab:bbox_region_counts}), and grounding for these regions remained substantially weaker than for frequently annotated anatomy even after ACCO (Fig.~\ref{fig:cxr_evaluation_composite}g). Because the lung zones and apical regions provide parenchymal and pleural-boundary evidence relevant to Consolidation and Pneumothorax, respectively, weak grounding may contribute to ARIVE's low CheXbert $\mathrm{F}_{1}$ for these observations (Fig.~\ref{fig:cxr_evaluation_composite}a,d), although this explanation remains hypothetical. Second, the anatomical boxes are predefined regions rather than lesions, image slices or volumetric structures. Extending ARIVE beyond chest radiography will therefore require modality-specific spatial supervision, trajectory design and evaluation.  Third, we used only frontal radiographs and did not explore integrating lateral views or other complementary imaging information into the diagnostic trajectory. Fourth, because the MCoT trajectories were reconstructed from radiology reports by a large language model, they may inherit omissions and institution-specific reporting practices from the source reports, as well as errors introduced during rewriting. They should therefore not be interpreted as faithful reconstructions of radiologist cognition. Finally, the clinical evaluation datasets derive from MIMIC-CXR, a single-institution resource, and generalization to other institutions and reporting styles remains to be established.

Within these limitations, ARIVE shows that anatomical inspection can be integrated into report generation to explicitly link regional visual evidence, intermediate interpretations and final findings. This approach may support more inspectable reporting, although its applicability beyond chest radiography requires evaluation. Future work will examine how models can learn to identify diagnostically relevant regions and predict their locations without bounding-box annotations during training, which would reduce reliance on modality-specific spatial annotations and facilitate extension to other imaging modalities.
\FloatBarrier

\section{Methods}
\label{sec:methods}

\subsection{Ethics statement}
\label{sec:ethics}

This study used de-identified chest radiographs and reports from MIMIC-CXR and its MIMIC-ABN subset~\cite{ni2020learning}, together with anatomical annotations from Chest ImaGenome~\cite{wu2021chest}. The
creation of MIMIC-CXR was reviewed by the Institutional Review Board of Beth Israel Deaconess Medical Center, which granted a waiver of informed consent and approved the data sharing initiative~\cite{johnson2019mimic}. The present study was a secondary analysis of de-identified datasets and involved no new patient recruitment or collection of identifiable patient information. All data were handled in accordance with the data use agreements and licensing requirements of each dataset.

All models that processed these data, including GPT-OSS-120B,
DeepSeek-V4.1-Flash and Qwen3.5-397B-A17B, were run on local
infrastructure, and no data were sent to external online services.

\subsection{Data sources}
\label{sec:datasets}
MIMIC-CXR~\cite{johnson2019mimic} is a publicly available, de-identified chest radiograph dataset comprising 377,110 radiographs from more than 227,000 studies, each study accompanied by a free-text report. MIMIC-ABN~\cite{ni2020learning} is an abnormality-centered subset of MIMIC-CXR that includes only abnormal studies with reports restricted to abnormality-related findings. Chest ImaGenome~\cite{wu2021chest} provides anatomy-centered scene graphs for MIMIC-CXR, linking anatomical bounding boxes to clinical attributes. It includes automatically generated Silver annotations covering 29 anatomical regions and a manually annotated Gold subset. 

Following prior work~\cite{zhang2025libra,liu2025enhanced,liu2026priorrg}, we used only the Findings section as the reference report. For both datasets, we retained frontal anteroposterior and posteroanterior radiographs, excluded lateral and other non-frontal views, and followed the official data splits. To obtain the anatomical supervision required for MCoT construction, we further restricted each training split to studies represented in the Chest ImaGenome Silver subset. Dataset statistics after these selection criteria are reported in Supplementary Table~\ref{supp-tab:dataset_statistics}, and CheXbert positive-label counts for each split are provided in Supplementary Table~\ref{supp-tab:CheXbert_label_counts}. 
Anatomical grounding was evaluated using the Chest ImaGenome Gold samples associated with each dataset. The MIMIC-CXR Gold subset comprised 636 samples (from 636 studies of 444 patients). The MIMIC-ABN Gold subset comprised 142 samples (from 142 studies of 120 patients). Because the Chest ImaGenome Gold subset does not provide bounding boxes for the spine or abdomen, these regions were excluded from the grounding evaluation.

\FloatBarrier

\subsection{Construction of the multimodal chain-of-thought dataset}
\label{sec:mcot_dataset}
We constructed the MCoT dataset to teach ARIVE an explicit diagnostic trajectory rather than only a direct image-to-report mapping. Starting from the filtered MIMIC-CXR training set, we generated 159,416 MCoT trajectories by integrating two complementary sources~\cite{myronenko2025reasoning,wu2021chest}: radiologist-style reasoning traces, which provided diagnostic structure and interpretation order, and Chest ImaGenome Silver annotations, which provided bounding boxes for 29 anatomical regions.

We first used radiologist-style reasoning traces released by NV-Reason-CXR~\cite{myronenko2025reasoning} as diagnostic templates. These traces organize chest radiograph interpretation into a standardized sequence comprising device inspection, airway evaluation, bilateral lung examination, mediastinal and hilar assessment, cardiac assessment, upper-abdominal inspection, bone review, and diagnostic summarization. This sequence provided a structure for reorganizing report findings into stepwise diagnostic interpretations. Using these templates, we prompted GPT-OSS-120B~\cite{openai2025gptoss} to rewrite each source report as a textual CoT trajectory. The prompt instructed the model to preserve all diagnostic information while reorganizing the report content into a stepwise account of diagnostic reasoning. 

Second, we incorporated image-specific anatomical bounding boxes from Chest ImaGenome~\cite{wu2021chest} to convert textual CoT into MCoT. Chest ImaGenome provides bounding-box annotations for 29 anatomical regions in MIMIC-CXR radiographs. These image-specific annotations allowed us to insert the corresponding bounding boxes into clinically significant reasoning steps. For each textual reasoning trajectory, we prompted GPT-OSS-120B to insert anatomical grounding only for clinically significant steps whose referenced anatomy was mentioned in the source report and corresponded to one of the 29 predefined Chest ImaGenome anatomical regions. This selective grounding strategy restricted region requests to report-relevant anatomy and retained steps without applicable regional annotations as text-only reasoning. Each completed MCoT trajectory was paired with its corresponding final report, jointly supervising anatomical region selection, regional interpretation, and report composition within a unified generation sequence. The 159,416 training trajectories contained an average of 399.98 words of reasoning text and 5.75 region-grounding instances per trajectory (Supplementary Table~\ref{supp-tab:mcot_trajectory_statistics}).
The prompts for textual CoT construction and anatomical bounding-box insertion are provided in Supplementary Tables~\ref{supp-tab:prompt_textual_cot} and~\ref{supp-tab:prompt_bbox_insertion}.

\subsection{Interleaved anatomical inspection and diagnostic reasoning}
\label{sec:interleaved_reasoning}
\begin{figure*}[!t]
\centering
\includegraphics[width=0.88\textwidth]{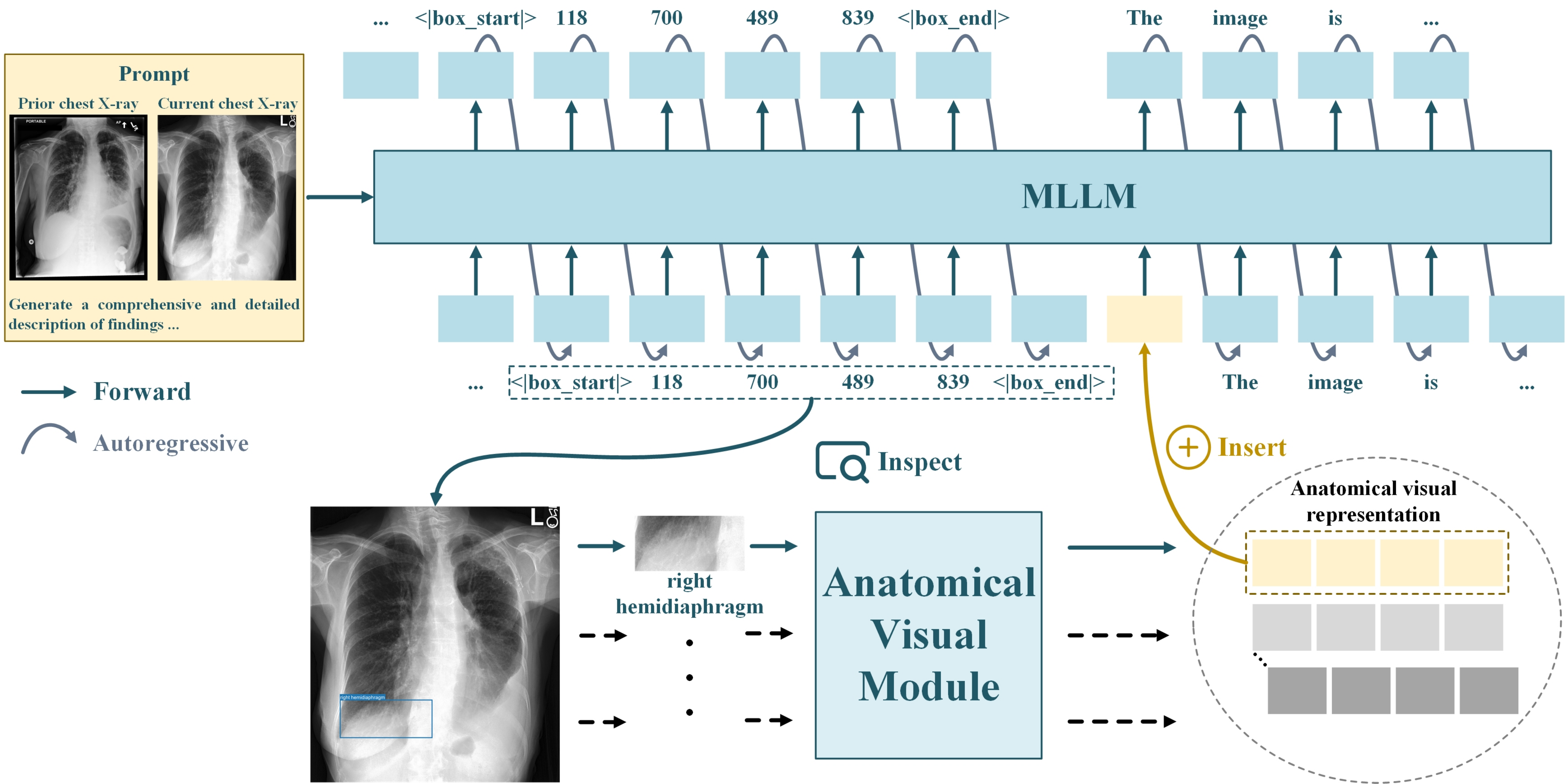}
\caption{Interleaved anatomical inspection and diagnostic reasoning in ARIVE. During generation, ARIVE inspects selected anatomical regions and inserts their visual representations into the model context to guide subsequent reasoning.}
\label{fig:agent_inspection}
\end{figure*}

ARIVE augments the backbone MLLM with an anatomical visual module that encodes regional image evidence during reasoning (Fig.~\ref{fig:agent_inspection}). The backbone visual encoder captures the radiograph globally, whereas the anatomical visual module encodes local image crops specified by bounding boxes. Whenever a bounding box is generated or otherwise provided, the corresponding anatomical region is cropped from the radiograph and transformed into a local visual representation. This representation is then inserted into the MLLM context before generation resumes, allowing subsequent interpretation to incorporate the original prompt, the global image representation, the accumulated diagnostic text, and the newly inspected regional evidence.

During inference, successive region requests establish an iterative sequence of regional inspection and interpretation. After receiving each local visual representation, the MLLM continues generating diagnostic text and either requests an additional anatomical region or completes the trajectory. Once the reasoning trajectory is complete, the model generates the final report conditioned on the accumulated global and regional visual evidence with the preceding diagnostic context. Repeated regional inspection makes localized visual evidence available during generation. Recording each region request and its associated interpretation allows reviewers to inspect the links between regional evidence, intermediate statements and the final report. 

\subsection{Model training}
\label{sec:model_training}

\subsubsection{Supervised fine-tuning}
\label{sec:sft}
We first trained ARIVE on the MCoT dataset with SFT. Each training example is represented by the tuple $(X,R,Y)$, where $X=(\mathcal{I},P)$ comprises the input radiographs and textual prompt. The ordered image tuple is $\mathcal{I}=(I^{\mathrm{prior}},I^{\mathrm{cur}})$ when the most recent prior radiograph is available, and $\mathcal{I}=(I^{\mathrm{cur}})$ otherwise. Here, $I^{\mathrm{prior}}$ denotes the most recent prior radiograph, and $I^{\mathrm{cur}}$ denotes the current radiograph. The prompt $P$ contains the task instructions, the available \textit{Indication} text, and the prior report when available. The structured MCoT trajectory $R$ specifies region requests, diagnostic interpretations, and where local visual representations are inserted and $Y$ denotes the final report.

We serialized $R$ followed by $Y$ into an interleaved multimodal sequence $S=(s_1,s_2,\ldots,s_T)$, inserting encoded local visual representations at the specified positions. Each $s_t$ represents a textual token or a local visual embedding, so a visual block may occupy multiple sequence positions. At position $t$, $X$ provides the initial input context, and $S_{<t}=(s_1,\ldots,s_{t-1})$ contains the preceding text and local visual embeddings.

We optimized the model with an autoregressive language-modeling objective over textual prediction targets. Let $m_t=1$ for textual tokens to be predicted and $m_t=0$ for externally encoded local visual embeddings. With $\mathcal{T}_{\mathrm{text}}=\{t\in\{1,\ldots,T\}:m_t=1\}$, the SFT loss is
\begin{equation}
\mathcal{L}_{\mathrm{SFT}}
= -\sum_{t\in\mathcal{T}_{\mathrm{text}}} \log p_{\theta}\!\left(s_t \mid X,S_{<t}\right).
\label{eq:sft_loss}
\end{equation}
Local visual embeddings remain available in $S_{<t}$ for subsequent predictions but are excluded from the prediction targets. The loss is therefore computed only at textual positions.

\subsubsection{Anatomical and Clinical-Content Optimization}
\label{sec:lado}
SFT trains ARIVE to generate interleaved MCoT trajectories, but its token-level objective does not directly reward anatomical grounding accuracy or agreement with target clinical content. ACCO uses group relative policy optimization (GRPO)~\cite{guo2025deepseek} to optimize generated trajectories and reports through four reward components. To account for imbalanced observation and anatomical-region frequencies, inverse-frequency weights are applied to the bounding-box and CheXbert rewards, while the RadGraph evaluates entity--relation structure without frequency weighting.

Given $X=(\mathcal{I},P)$, the current policy samples textual tokens for a group of $N$ responses $\{S_i\}_{i=1}^{N}$, each comprising an MCoT trajectory and a final report. During generation, the anatomical visual module inserts local visual embeddings in response to region requests, yielding the interleaved sequence $S_i=(s_{i,1},\ldots,s_{i,T_i})$. These embeddings provide context for subsequent token sampling. For each sampled response $i$, we compute four rewards: anatomical grounding $R_{i,\mathrm{BBOX}}$, reasoning-stage CheXbert $R^{\mathrm{rea}}_{i,\mathrm{CXB}}$, final-report CheXbert $R^{\mathrm{ans}}_{i,\mathrm{CXB}}$, and final-report RadGraph $R_{i,\mathrm{RG}}$. Because their numerical scales differ, each component is normalized within the sampled group:
\begin{align}
\widetilde{R}_{i,k} &= \frac{R_{i,k}-\mu_k}{\sigma_k+\epsilon_r}, \nonumber\\
\mu_k &= \frac{1}{N}\sum_{i=1}^{N}R_{i,k}, \nonumber\\
\sigma_k &= \sqrt{\frac{1}{N}\sum_{i=1}^{N}\left(R_{i,k}-\mu_k\right)^2},
\label{eq:reward_normalization}
\end{align}
where $k\in\{\mathrm{BBOX},\mathrm{CXB}^{\mathrm{rea}},\mathrm{CXB}^{\mathrm{ans}},\mathrm{RG}\}$ and $\epsilon_r$ is a small constant for numerical stability.

The normalized components are combined into a sequence-level advantage:
\begin{equation}
A_i = \frac{1}{4}
\left(\widetilde{R}_{i,\mathrm{BBOX}}+\widetilde{R}^{\mathrm{rea}}_{i,\mathrm{CXB}}
+
\widetilde{R}^{\mathrm{ans}}_{i,\mathrm{CXB}}+\widetilde{R}_{i,\mathrm{RG}}\right).
\label{eq:sequence_advantage}
\end{equation}

At the token level, ACCO follows GRPO and optimizes a clipped policy objective with Kullback–Leibler (KL) regularization. For response $i$, let $\mathcal{T}_{i,\mathrm{text}}=\{t\in\{1,\ldots,T_i\}:m_{i,t}=1\}$ denote the textual prediction positions, using the same mask definition as in SFT. The objective is
\begin{align}
\mathcal{J}_{\mathrm{ACCO}}(\theta)
&= \frac{1}{N}\sum_{i=1}^{N}\frac{1}{M_i}\sum_{t\in\mathcal{T}_{i,\mathrm{text}}}
\min\!\left(r_{i,t}(\theta)A_i,\overline{r}_{i,t}(\theta)A_i\right) \nonumber\\
&\quad -\eta D_{\mathrm{KL}}\!\left(\pi_{\theta}(\cdot\mid X)\,\|\,\pi_{\mathrm{ref}}(\cdot\mid X)\right),
\label{eq:lado_objective}
\end{align}
with the probability ratios defined for $t\in\mathcal{T}_{i,\mathrm{text}}$ as
\begin{align}
r_{i,t}(\theta)
&= \frac{\pi_{\theta}(s_{i,t}\mid X,S_{i,<t})}
{\pi_{\theta_{\mathrm{old}}}(s_{i,t}\mid X,S_{i,<t})}, \nonumber\\
\overline{r}_{i,t}(\theta)
&= \operatorname{clip}\!\left(r_{i,t}(\theta),1-\epsilon,1+\epsilon\right).
\label{eq:policy_ratio}
\end{align}
Here, $M_i=|\mathcal{T}_{i,\mathrm{text}}|=\sum_{t=1}^{T_i}m_{i,t}$ is the number of textual prediction targets. The terms $r_{i,t}(\theta)$ and $\overline{r}_{i,t}(\theta)$ denote the token-level probability ratio and its clipped value, respectively. The clipping threshold is $\epsilon$, and $\eta$ controls KL regularization toward the reference policy $\pi_{\mathrm{ref}}$. As in SFT, local visual embeddings contribute to the conditioning context $S_{i,<t}$ but are excluded from the token-level prediction terms.
\begingroup
\setlength{\abovedisplayskip}{6pt plus 2pt minus 1pt}
\setlength{\belowdisplayskip}{6pt plus 2pt minus 1pt}
\setlength{\jot}{2pt}
\paragraph{Weighted bounding-box reward.}
This reward evaluates the selection and localization of the anatomical regions specified by the target MCoT trajectory. For each anatomical category $c$, its empirical training frequency is
\begin{equation}
p_c = \frac{n_c}{\sum_j n_j},
\label{eq:region_frequency}
\end{equation}
where $n_c$ is the number of training annotations for category $c$. To reduce the influence of frequently annotated regions, we used the normalized inverse-frequency weight
\begin{equation}
w_c = \frac{p_c^{-\alpha_{\mathrm{reg}}}}
{\frac{1}{|\mathcal{C}|}\sum_{j\in\mathcal{C}}p_j^{-\alpha_{\mathrm{reg}}}},
\label{eq:region_weight}
\end{equation}
where $\mathcal{C}$ is the set of anatomical categories and $\alpha_{\mathrm{reg}}$ controls the strength of region reweighting.

Let $\mathcal{G}$ and $\mathcal{P}$ denote the ground-truth and predicted region sets. For each $c\in\mathcal{G}$, the predicted box $\hat{b}_c$ is matched to the ground-truth box $b_c$ from the same category. If category $c$ is not predicted, its intersection over union (IoU) is set to zero. The reward is
\begin{equation}
R_{\mathrm{BBOX}}
= \frac{\sum_{c\in\mathcal{G}}w_c\,\operatorname{IoU}(b_c,\hat{b}_c)}
{\sum_{c\in\mathcal{G}}w_c}
-\lambda\max\!\left(0,|\mathcal{P}|-|\mathcal{G}|\right),
\label{eq:bbox_reward}
\end{equation}
where $\lambda$ penalizes predictions beyond the ground-truth count. Low overlap, missing categories and excess requests reduce the reward, while $w_c$ increases the contribution of underrepresented regions.

\paragraph{Weighted reasoning CheXbert reward.}
This reward evaluates observation-level correctness in intermediate diagnostic text. A CheXbert-style classifier adapted to reasoning trajectories extracts 14 observation labels~\cite{smit2020combining}. Its outputs for generated reasoning are denoted by $\hat{y}$, and $y$ denotes the target labels.

Let $\mathcal{O}$ be the set of CheXbert observations and $n_o^{+}$ the number of positive training labels for observation $o$. Its positive-label frequency is $q_o=n_o^{+}/\sum_{j\in\mathcal{O}}n_j^{+}$. We obtain $v_o$ from Eq.~\eqref{eq:region_weight} by substituting $q_o$, $\mathcal{O}$ and $\alpha_{\mathrm{obs}}$ for $p_c$, $\mathcal{C}$ and $\alpha_{\mathrm{reg}}$, respectively. The weighted counts and corresponding $\mathrm{F}_{1}$ reward are
\begin{equation}
\begin{aligned}
\mathrm{TP}_{w} &= \sum_{o\in\mathcal{O}} v_o\,\mathbf{1}[\hat{y}_o=1\wedge y_o=1],\\
\mathrm{FP}_{w} &= \sum_{o\in\mathcal{O}} v_o\,\mathbf{1}[\hat{y}_o=1\wedge y_o=0],\\
\mathrm{FN}_{w} &= \sum_{o\in\mathcal{O}} v_o\,\mathbf{1}[\hat{y}_o=0\wedge y_o=1],\\
R^{\mathrm{rea}}_{\mathrm{CXB}}
&= \frac{2\mathrm{TP}_{w}}
{2\mathrm{TP}_{w}+\mathrm{FP}_{w}+\mathrm{FN}_{w}}.
\end{aligned}
\label{eq:reasoning_chexbert_reward}
\end{equation}
Here, TP, FP and FN denote true positives, false positives and false negatives, respectively, and the subscript $w$ indicates weighted counts. Because $v_o$ increases as positive-label frequency decreases, observations with fewer positive examples contribute more strongly.
\endgroup
\paragraph{Weighted final-report CheXbert reward.}
This reward applies the same observation weights and weighted $\mathrm{F}_{1}$ to CheXbert labels extracted from the generated and reference reports. Substituting these report-level label vectors into Eq.~\eqref{eq:reasoning_chexbert_reward} gives $R^{\mathrm{ans}}_{\mathrm{CXB}}$, which emphasizes clinically correct final-report observations, including underrepresented categories.

\paragraph{RadGraph reward.}
The CheXbert rewards assess clinical efficacy over a fixed set of observations. We added a RadGraph reward to encourage the final report to preserve clinical entities and their relationships in the reference report. For each generated report, we computed the Simple, Partial and Complete RadGraph scores~\cite{delbrouck2022improving} against its reference. Simple matches entities by text and label, whereas Partial additionally checks for outgoing relations. Complete further checks relation types and target entities.  We define $R_{\mathrm{RG}}$ as the mean of these three scores.
\begin{equation}
R_{\mathrm{RG}}
= \frac{1}{3}\left(\mathrm{RG}_{\mathrm{simple}}+\mathrm{RG}_{\mathrm{partial}}+\mathrm{RG}_{\mathrm{complete}}\right).
\label{eq:radgraph_reward}
\end{equation}

Together, the four rewards supervise anatomical grounding, intermediate and final-report observation content, and final-report entity--relation structure. Frequency weighting increases contributions from underrepresented anatomical regions and clinical observations.

\subsection{Implementation details}
\label{sec:implementation_details}
Following recent prior-aware RRG studies~\cite{liu2026priorrg,zhang2025libra,hou2025radar}, we incorporated available patient-specific historical information into the model input. For each examination, the model received the current radiograph together with the most recent prior radiograph and its corresponding report, when available. For the current examination's clinical context, we retain only the \textit{Indication} section. This setting enables the model to leverage longitudinal imaging history and relevant clinical cues, thereby more closely approximating the information available during real-world radiological interpretation. The report-generation prompts for MIMIC-CXR and MIMIC-ABN are provided in Supplementary Tables~\ref{supp-tab:prompt_mimic_cxr} and~\ref{supp-tab:prompt_mimic_abn}, respectively. Considering both model generalization and training cost, we used Qwen3-VL-Instruct-4B~\cite{bai2025qwen3} as the MLLM backbone. The anatomical visual module mirrored the architecture of the backbone visual module, including its vision encoder and visual merger, and was initialized with the corresponding backbone parameters. All input images were resized to $224\times224$ pixels (50,176 pixels).

During supervised fine-tuning (SFT), we jointly optimized the backbone visual module, anatomical visual module, and LLM using AdamW with a learning rate of $1\times10^{-5}$. Training used a batch size of 64, cosine learning-rate decay, and a warmup ratio of 0.03. SFT on MIMIC-CXR was performed for six epochs. The MIMIC-ABN model was subsequently initialized from the resulting MIMIC-CXR checkpoint and further adapted for three epochs on the MIMIC-ABN training split. During ACCO, we updated only the LLM for one epoch using AdamW with a learning rate of $1\times10^{-6}$, a batch size of 32, and a group size of five. The KL-divergence coefficient was set to 0.01. The bounding-box penalty coefficient $\lambda$ was set to 0.05. We used inverse-frequency exponents of $\alpha_{\mathrm{reg}}=0.25$ for anatomical regions and $\alpha_{\mathrm{obs}}=0.5$ for CheXbert observations. These hyperparameters were selected empirically within the available computational budget to maintain comparable numerical contributions across the different reward sources. Inference was accelerated using vLLM 0.19.0~\cite{kwon2023efficient}.

\subsection{Comparison baselines and protocols}
\label{sec:baselines}
We compared ARIVE with three groups of baseline models: a general-purpose MLLM, a generalist medical MLLM, and task-specific RRG models.
Qwen3.5-397B-A17B~\cite{qwen3.5} is an open-weight MLLM with a sparse mixture-of-experts architecture comprising 397 billion total parameters and 17 billion active parameters, evaluated here in reasoning mode.
MedGemma-1.5~\cite{sellergren2026medgemma} is a generalist medical MLLM supporting medical image and text understanding, including localization and longitudinal chest X-ray analysis.

The task-specific RRG baselines comprised MAIRA-2, Libra, MLRG, and PriorRG.
MAIRA-2~\cite{bannur2024maira} couples a frozen RAD-DINO image encoder with a Vicuna-7B language model through a four-layer multilayer perceptron adapter.
Libra~\cite{zhang2025libra} connects a RAD-DINO image encoder to Meditron-7B through a Temporal Alignment Connector that integrates visual features from current and prior radiographs.
MLRG~\cite{liu2025enhanced} combines RAD-DINO and CXR-BERT encoders with multi-view longitudinal and multimodal fusion networks to condition a DistilGPT2 report decoder.
PriorRG~\cite{liu2026priorrg} uses RAD-DINO and CXR-BERT encoders with a DistilGPT2 decoder, progressively integrating clinical context, spatiotemporal features and hierarchical visual features through coarse-to-fine fusion.

Whenever available, we used publicly released checkpoints or official model-access configurations. All models received the same case-level input content as ARIVE and were evaluated on the same test sets using the same metric implementations. For the task-specific baselines MAIRA-2 and Libra, which were not originally trained on MIMIC-ABN, we followed their published training protocols by initializing each model from its publicly released checkpoint and further training it on the MIMIC-ABN training split before evaluation.

We also compared model size and inference time on the MIMIC-CXR test set (Supplementary Table~\ref{supp-tab:mimic_inference_profile}). With 4.85 billion parameters, ARIVE generated an MCoT trajectory and a final report in a mean of 1.13 s per sample on a single NVIDIA A100 GPU, close to the report-only inference times of MedGemma-1.5 and MAIRA-2. ARIVE used fewer parameters than MAIRA-2 and Libra, whereas the smaller MLRG and PriorRG models were faster.

\subsection{Evaluation metrics}
\label{sec:evaluation_metrics}
Clinical-efficacy metrics included macro- and micro-averaged CheXbert $\mathrm{F}_{1}$~\cite{smit2020combining} over 14 CheXpert observations~\cite{irvin2019chexpert}, together with five-observation CheXbert $\mathrm{F}_{1}$ over Atelectasis, Cardiomegaly, Consolidation, Edema, and Pleural Effusion. We additionally reported $\mathrm{F}_{1}\text{-}\mathrm{SRR}\text{-}\mathrm{BERT}$~\cite{delbrouck2025automated} over 55 radiology labels.
RadGraph $\mathrm{F}_{1}$~\cite{jain2021radgraph} evaluated the clinical entity--relation structure of generated reports. Entities were matched by text and label, and relations by source and target entities, type and direction. The score for each generated--reference report pair was the mean of the entity and relation $\mathrm{F}_{1}$ values:
\begin{equation}
\mathrm{RadGraph}\ \mathrm{F}_{1}
= \frac{1}{2}\left(\mathrm{F}_{1,\mathrm{entity}}+\mathrm{F}_{1,\mathrm{relation}}\right).
\label{eq:radgraph_evaluation}
\end{equation}
This evaluation score differs from the training reward $R_{\mathrm{RG}}$, which averages the RadGraph Simple, Partial and Complete scores (Eq.~\eqref{eq:radgraph_reward}).

GREEN~\cite{ostmeier2024green} evaluated matched findings and clinically significant errors. For each generated--reference report pair, the score was
\begin{equation}
\mathrm{GREEN}
= \frac{N_{\mathrm{match}}}{N_{\mathrm{match}}+N_{\mathrm{error}}},
\label{eq:green_evaluation}
\end{equation}
where $N_{\mathrm{match}}$ is the number of matched findings and $N_{\mathrm{error}}$ is the total number of clinically significant errors across GREEN's six error categories. The score was set to zero when $N_{\mathrm{match}}=0$. GREEN ranges from 0 to 1, with higher values indicating better report quality. We did not assess GREEN on MIMIC-ABN because its reports were restricted to abnormality-related findings.
Lexical similarity was evaluated using BLEU-4~\cite{papineni2002bleu}, METEOR~\cite{banerjee2005meteor}, and ROUGE-L~\cite{lin2004rouge}. These metrics capture higher-order n-gram precision, unigram-level semantic correspondence, and longest-common-subsequence overlap, respectively.

For MCoT trajectories, region-selection $\mathrm{F}_{1}$ evaluated the selection of reference anatomical regions for local inspection, whereas mean intersection over union (mIoU) measured localization overlap averaged across all ground-truth anatomical regions. Reasoning-stage CheXbert evaluated the clinical content expressed during intermediate reasoning, and reasoning-report consistency of CheXbert quantified agreement between the clinical content expressed in the reasoning trajectory and those in the final report. Training details for reasoning CheXbert are described below, and held-out performance is reported in Supplementary Table~\ref{supp-tab:reasoning_chexbert_metrics}.
Following the training protocol of the original CheXbert study~\cite{smit2020combining}, we trained reasoning CheXbert on the MCoT training data. We randomly partitioned the data at the patient level into 70\% training and 30\% held-out subsets, ensuring that all examples from the same patient remained in a single subset. We initialized the classifier with CheXbert parameters and fine-tuned it for 8 epochs to adapt label extraction from conventional radiology reports to reasoning-style diagnostic trajectories.
For the held-out evaluation of this classifier, uncertain labels were treated as positive.

\subsection{Statistical analysis}
\label{sec:statistics}
We used one-sided nonparametric tests to evaluate the directional hypothesis that ARIVE achieved higher scores than the baseline models. For corpus-level metrics, including CheXbert $\mathrm{F}_{1}$, $\mathrm{F}_{1}\text{-}\mathrm{SRR}\text{-}\mathrm{BERT}$, and BLEU-4, we used paired permutation tests with resampling clustered at the patient level. For report-level metrics, including METEOR, ROUGE-L, GREEN, and RadGraph $\mathrm{F}_{1}$, we used paired Wilcoxon signed-rank tests. Two-sided 95\% confidence intervals (CIs) were reported to describe estimation uncertainty. For each model, CIs were estimated from 10,000 patient-clustered bootstrap resamples. For pairwise model comparisons, bootstrap resampling was paired by applying the same patient-level resample to both models. CI bounds were defined by the 2.5th and 97.5th percentiles of the resulting bootstrap distribution.

\FloatBarrier

\backmatter

\section*{Data availability}
The MIMIC-CXR v2.1.0 (\url{https://doi.org/10.13026/4jqj-jw95}) and Chest ImaGenome v1.0.0 (\url{https://doi.org/10.13026/wv01-y230}) datasets are available on PhysioNet~\cite{pollard_physionet_2026} after a credentialing process. MIMIC-ABN is an abnormality-centered subset of MIMIC-CXR, with the preprocessing procedure adopted in this study detailed in \url{https://doi.org/10.18653/v1/2020.findings-emnlp.176}.

\section*{Code availability}
The code for MCoT construction and ARIVE training and inference is available at \url{https://github.com/DUTIR-BioNLP/ARIVE}.

\setlength{\bibsep}{0pt plus 1pt}
\bibliography{references}

\begingroup
\makeatletter
\renewcommand\section{\@startsection{section}{1}{\z@}%
  {-8pt plus -2pt minus -1pt}{6pt}{\sectionfont}}
\makeatother
\section*{Funding}
This research was supported by the National Natural Science Foundation of China (No. 62676065) and the Fundamental Research Funds for the Central Universities (No. DUT25YG108).

\section*{Author contributions}
C.H. and L.L. conceived this work. C.H. designed the framework and conducted the experiments. Z.W. and Y.L. contributed to the data acquisition and provided suggestions on the framework. R.L. performed the clinical evaluation. L.L. polished the manuscript and provided suggestions on the experiments. Y.S., J.W., H.L. and L.L. supervised the project. All authors discussed the results and contributed to the final manuscript.

\section*{Competing interests}
The authors declare no competing interests.
\endgroup

\clearpage
\appendix
\input{supplementary_information_body}

\end{document}

%% file: supplementary_information_body.tex
\begingroup
\setcounter{section}{0}
\setcounter{subsection}{0}
\setcounter{table}{0}
\renewcommand{\thesection}{S\arabic{section}}
\renewcommand{\theHsection}{supp.\arabic{section}}
\renewcommand{\tablename}{Supplementary Table}
\renewcommand{\thetable}{S\arabic{table}}
\renewcommand{\theHtable}{supp.\arabic{table}}
\setlength{\RRGNLGLayoutWidth}{1.4103\textwidth}
\renewcommand{\floatpagefraction}{0.75}
\renewcommand{\dblfloatpagefraction}{0.75}
\pdfbookmark[1]{Appendix: Supplementary information}{app:supplementary_information}

\begin{center}
{\fontsize{17}{20}\selectfont Appendix: Supplementary Information\par}
\vspace{8pt}
{\small Interleaving anatomical inspection and diagnostic reasoning for explainable radiology report generation\par}
\end{center}
\vspace{4pt}
\setlength{\intextsep}{6pt plus 2pt minus 2pt}

\section{Natural language generation results}

\begin{table*}[!ht]
\centering
\caption{Natural language generation performance of ARIVE and comparator models on MIMIC-CXR and MIMIC-ABN. Values are reported with 95\% CIs in parentheses. Bold marks the highest value for each metric within each dataset. Stars indicate ARIVE's superiority in one-sided comparisons using patient-clustered paired permutation tests for BLEU-4 and paired Wilcoxon signed-rank tests for METEOR and ROUGE-L. $^{*}p < 0.05$; $^{**}p < 0.01$.}
\label{supp-tab:full_report_nlg_metrics}
\begingroup
\renewcommand{\arraystretch}{1.80}
\setlength{\tabcolsep}{2.5pt}
\settowidth{\RRGNLGFixedWidth}{\mbox{MIMIC-CXR}\mbox{Qwen3.5-397B-A17B}}
\begin{adjustbox}{width=\textwidth}
\begin{tabular}{cl*{3}{>{\centering\arraybackslash}p{\dimexpr(\RRGNLGLayoutWidth-\RRGNLGFixedWidth-10\tabcolsep)/3\relax}}}
\toprule
\rowcolor{TableHeader}
\textbf{Dataset} & \textbf{Model}
& BLEU-4 & METEOR & ROUGE-L \\
\midrule
\rowcolor{TableRowBlue}
& Qwen3.5-397B-A17B & \metricci{0.062\sigmark{**}}{0.058}{0.065} & \metricci{0.344}{0.338}{0.349} & \metricci{0.202\sigmark{**}}{0.198}{0.206} \\
\cmidrule[0.35pt]{2-5}
\rowcolor{TableRowGrey}
& MedGemma-1.5 & \metricci{0.082\sigmark{**}}{0.077}{0.087} & \metricci{0.311\sigmark{**}}{0.305}{0.318} & \metricci{0.230\sigmark{**}}{0.225}{0.235} \\
\cmidrule[0.35pt]{2-5}
\rowcolor{TableRowBlue}
& MAIRA-2 & \metricci{0.098\sigmark{**}}{0.093}{0.103} & \metricci{0.268\sigmark{**}}{0.262}{0.275} & \metricci{0.238\sigmark{**}}{0.232}{0.243} \\
\cmidrule[0.35pt]{2-5}
\rowcolor{TableRowGrey}
& Libra & \metricci{0.133\sigmark{*}}{0.126}{0.140} & \metricci{0.327}{0.320}{0.334} & \metricci{0.253\sigmark{**}}{0.247}{0.260} \\
\cmidrule[0.35pt]{2-5}
\rowcolor{TableRowBlue}
& MLRG & \metricci{0.150}{0.143}{0.157} & \metricci{0.324\sigmark{**}}{0.316}{0.332} & \metricci{0.294}{0.286}{0.301} \\
\cmidrule[0.35pt]{2-5}
\rowcolor{TableRowGrey}
& PriorRG & \metricci{\textbf{0.175}}{0.168}{0.182} & \metricci{\textbf{0.355}}{0.347}{0.363} & \metricci{\textbf{0.318}}{0.310}{0.325} \\
\cmidrule[0.35pt]{2-5}
\rowcolor{TableARIVE}
\multirow{-7}{*}{MIMIC-CXR} & ARIVE (ours) & \metricci{0.139}{0.133}{0.145} & \metricci{0.328}{0.321}{0.335} & \metricci{0.284}{0.279}{0.290} \\
\midrule
\rowcolor{TableRowBlue}
& Qwen3.5-397B-A17B & \metricci{0.055\sigmark{**}}{0.047}{0.062} & \metricci{\textbf{0.285}}{0.272}{0.297} & \metricci{0.180\sigmark{**}}{0.170}{0.188} \\
\cmidrule[0.35pt]{2-5}
\rowcolor{TableRowGrey}
& MedGemma-1.5 & \metricci{0.052\sigmark{**}}{0.043}{0.061} & \metricci{0.256}{0.243}{0.269} & \metricci{0.186\sigmark{**}}{0.175}{0.196} \\
\cmidrule[0.35pt]{2-5}
\rowcolor{TableRowBlue}
& MAIRA-2 & \metricci{0.067\sigmark{**}}{0.058}{0.076} & \metricci{0.214\sigmark{**}}{0.202}{0.228} & \metricci{0.190\sigmark{**}}{0.179}{0.202} \\
\cmidrule[0.35pt]{2-5}
\rowcolor{TableRowGrey}
& Libra & \metricci{0.056\sigmark{**}}{0.049}{0.063} & \metricci{0.279}{0.267}{0.290} & \metricci{0.181\sigmark{**}}{0.171}{0.192} \\
\cmidrule[0.35pt]{2-5}
\rowcolor{TableRowBlue}
& MLRG & \metricci{0.101}{0.084}{0.121} & \metricci{0.261}{0.242}{0.282} & \metricci{\textbf{0.228}}{0.210}{0.248} \\
\cmidrule[0.35pt]{2-5}
\rowcolor{TableRowGrey}
& PriorRG & \metricci{\textbf{0.112}}{0.097}{0.129} & \metricci{0.264}{0.244}{0.284} & \metricci{\textbf{0.228}}{0.212}{0.244} \\
\cmidrule[0.35pt]{2-5}
\rowcolor{TableARIVE}
\multirow{-7}{*}{MIMIC-ABN} & ARIVE (ours) & \metricci{0.085}{0.074}{0.096} & \metricci{0.253}{0.239}{0.268} & \metricci{0.224}{0.211}{0.236} \\
\bottomrule
\end{tabular}
\end{adjustbox}
\endgroup
\end{table*}
\FloatBarrier

\begin{table*}[!ht]
\centering
\caption{Natural language generation performance across training settings on MIMIC-CXR and MIMIC-ABN. SFT uses report-only supervision; SFT w/ CoT and SFT w/ MCoT add text-only and region-interleaved diagnostic trajectories, respectively. ARIVE denotes SFT w/ MCoT followed by Anatomical and Clinical-Content Optimization (ACCO). Values are reported with 95\% CIs in parentheses. Bold marks the highest value for each metric within each dataset. Stars indicate ARIVE's superiority under the tests specified in Supplementary Table~\ref{supp-tab:full_report_nlg_metrics}. $^{*}p < 0.05$; $^{**}p < 0.01$.}
\label{supp-tab:full_training_ablation_nlg}
\begingroup
\renewcommand{\arraystretch}{1.80}
\setlength{\tabcolsep}{2.5pt}
\settowidth{\RRGNLGFixedWidth}{\mbox{MIMIC-CXR}\mbox{\textbf{Training setting}}}
\begin{adjustbox}{width=\textwidth}
\begin{tabular}{cl*{3}{>{\centering\arraybackslash}p{\dimexpr(\RRGNLGLayoutWidth-\RRGNLGFixedWidth-10\tabcolsep)/3\relax}}}
\toprule
\rowcolor{TableHeader}
\textbf{Dataset} & \textbf{Training setting}
& BLEU-4 & METEOR & ROUGE-L \\
\midrule
\rowcolor{TableRowBlue}
& SFT & \metricci{\textbf{0.149}}{0.143}{0.156} & \metricci{\textbf{0.330}}{0.325}{0.339} & \metricci{0.283\sigmark{*}}{0.277}{0.290} \\
\cmidrule[0.35pt]{2-5}
\rowcolor{TableRowGrey}
& SFT w/ CoT & \metricci{0.112\sigmark{**}}{0.107}{0.118} & \metricci{0.285\sigmark{**}}{0.278}{0.292} & \metricci{0.261\sigmark{**}}{0.256}{0.268} \\
\cmidrule[0.35pt]{2-5}
\rowcolor{TableRowBlue}
& SFT w/ MCoT & \metricci{0.123\sigmark{**}}{0.115}{0.127} & \metricci{0.303\sigmark{**}}{0.295}{0.309} & \metricci{0.267\sigmark{**}}{0.260}{0.272} \\
\cmidrule[0.35pt]{2-5}
\rowcolor{TableARIVE}
\multirow{-4}{*}{MIMIC-CXR} & ARIVE & \metricci{0.139}{0.133}{0.145} & \metricci{0.328}{0.321}{0.335} & \metricci{\textbf{0.284}}{0.279}{0.290} \\
\midrule
\rowcolor{TableRowBlue}
& SFT & \metricci{0.084}{0.072}{0.098} & \metricci{\textbf{0.253}}{0.237}{0.269} & \metricci{\textbf{0.228}}{0.215}{0.244} \\
\cmidrule[0.35pt]{2-5}
\rowcolor{TableRowGrey}
& SFT w/ CoT & \metricci{0.064\sigmark{**}}{0.063}{0.083} & \metricci{0.222\sigmark{**}}{0.217}{0.248} & \metricci{0.204\sigmark{*}}{0.199}{0.224} \\
\cmidrule[0.35pt]{2-5}
\rowcolor{TableRowBlue}
& SFT w/ MCoT & \metricci{0.075\sigmark{**}}{0.060}{0.079} & \metricci{0.239\sigmark{**}}{0.214}{0.242} & \metricci{0.218\sigmark{*}}{0.195}{0.220} \\
\cmidrule[0.35pt]{2-5}
\rowcolor{TableARIVE}
\multirow{-4}{*}{MIMIC-ABN} & ARIVE & \metricci{\textbf{0.085}}{0.074}{0.096} & \metricci{\textbf{0.253}}{0.239}{0.268} & \metricci{0.224}{0.211}{0.236} \\
\bottomrule
\end{tabular}
\end{adjustbox}
\endgroup
\end{table*}
\FloatBarrier

\Needspace{10\baselineskip}
\section{Dataset, annotation and trajectory statistics}
\label{supp-app:dataset_statistics}

\begin{table}[!ht]
\centering
\small
\renewcommand{\arraystretch}{1}
\caption{Sample, study and patient counts after frontal-view filtering and restriction of the training splits to Chest ImaGenome Silver overlap.}
\label{supp-tab:dataset_statistics}
\begin{tabular}{llrrr}
\toprule
\rowcolor{TableHeader}
\textbf{Dataset} & \textbf{Statistic} & \textbf{Train} & \textbf{Validation} & \textbf{Test} \\
\midrule
\rowcolor{TableRowBlue}
& Samples  & 159,416 & 1,286 & 2,463 \\
\cmidrule[0.35pt]{2-5}
\rowcolor{TableRowGrey}
& Studies  & 142,946 & 1,151 & 2,211 \\
\cmidrule[0.35pt]{2-5}
\rowcolor{TableRowBlue}
\multirow{-3}{*}{MIMIC-CXR}
& Patients & 57,729  & 448   & 285 \\
\midrule
\rowcolor{TableRowBlue}
& Samples  & 34,259 & 302 & 432 \\
\cmidrule[0.35pt]{2-5}
\rowcolor{TableRowGrey}
& Studies  & 34,152 & 263 & 378 \\
\cmidrule[0.35pt]{2-5}
\rowcolor{TableRowBlue}
\multirow{-3}{*}{MIMIC-ABN}
& Patients & 23,039 & 181 & 134 \\
\bottomrule
\end{tabular}
\end{table}
\FloatBarrier

\begingroup
\small
\setlength{\LTcapwidth}{\textwidth}
\setlength{\LTpre}{8pt}
\setlength{\LTpost}{10pt}
\begin{longtable}{lrrrr}
\caption{Anatomical bounding-box annotation counts in the training splits and Chest ImaGenome Gold subsets of MIMIC-CXR and MIMIC-ABN.}
\label{supp-tab:bbox_region_counts}\\
\toprule
\rowcolor{TableHeader}
& \multicolumn{2}{c}{\textbf{MIMIC-CXR}}
& \multicolumn{2}{c}{\textbf{MIMIC-ABN}} \\
\cmidrule(lr){2-3}\cmidrule(lr){4-5}
\rowcolor{TableHeader}
\multirow{-2}{*}{\textbf{Anatomical Region}}
& \multicolumn{1}{c}{\textbf{Train}} & \multicolumn{1}{c}{\textbf{Gold subset}}
& \multicolumn{1}{c}{\textbf{Train}} & \multicolumn{1}{c}{\textbf{Gold subset}} \\
\midrule
\endfirsthead
\multicolumn{5}{l}{\tablecaptionfont\textbf{Supplementary Table~\thetable} (continued)}\\
\toprule
\rowcolor{TableHeader}
& \multicolumn{2}{c}{\textbf{MIMIC-CXR}}
& \multicolumn{2}{c}{\textbf{MIMIC-ABN}} \\
\cmidrule(lr){2-3}\cmidrule(lr){4-5}
\rowcolor{TableHeader}
\multirow{-2}{*}{\textbf{Anatomical Region}}
& \multicolumn{1}{c}{\textbf{Train}} & \multicolumn{1}{c}{\textbf{Gold subset}}
& \multicolumn{1}{c}{\textbf{Train}} & \multicolumn{1}{c}{\textbf{Gold subset}} \\
\midrule
\endhead
\midrule
\multicolumn{5}{r}{\footnotesize\itshape Continued on next page}\\
\endfoot
\bottomrule
\endlastfoot
\rowcolor{TableRowBlue}
Right lung & 152,734 & 546 & 33,492 & 116 \\
\midrule[0.35pt]
\rowcolor{TableRowGrey}
Right upper lung zone & 4,757 & 46 & 1,015 & 12 \\
\midrule[0.35pt]
\rowcolor{TableRowBlue}
Right mid lung zone & 7,654 & 56 & 2,072 & 8 \\
\midrule[0.35pt]
\rowcolor{TableRowGrey}
Right lower lung zone & 31,442 & 174 & 6,706 & 35 \\
\midrule[0.35pt]
\rowcolor{TableRowBlue}
Right hilar structures & 48,140 & 343 & 11,108 & 81 \\
\midrule[0.35pt]
\rowcolor{TableRowGrey}
Right apical zone & 4,583 & 65 & 951 & 23 \\
\midrule[0.35pt]
\rowcolor{TableRowBlue}
Right costophrenic angle & 88,237 & 353 & 21,559 & 85 \\
\midrule[0.35pt]
\rowcolor{TableRowGrey}
Right hemidiaphragm & 8,905 & 80 & 2,577 & 23 \\
\midrule[0.35pt]
\rowcolor{TableRowBlue}
Left lung & 147,063 & 545 & 32,656 & 120 \\
\midrule[0.35pt]
\rowcolor{TableRowGrey}
Left upper lung zone & 4,204 & 32 & 673 & 7 \\
\midrule[0.35pt]
\rowcolor{TableRowBlue}
Left mid lung zone & 1,880 & 61 & 476 & 14 \\
\midrule[0.35pt]
\rowcolor{TableRowGrey}
Left lower lung zone & 33,104 & 190 & 6,928 & 43 \\
\midrule[0.35pt]
\rowcolor{TableRowBlue}
Left hilar structures & 5,970 & 352 & 1,268 & 88 \\
\midrule[0.35pt]
\rowcolor{TableRowGrey}
Left apical zone & 2,847 & 61 & 613 & 21 \\
\midrule[0.35pt]
\rowcolor{TableRowBlue}
Left costophrenic angle & 28,197 & 314 & 5,135 & 80 \\
\midrule[0.35pt]
\rowcolor{TableRowGrey}
Left hemidiaphragm & 3,157 & 51 & 644 & 8 \\
\midrule[0.35pt]
\rowcolor{TableRowBlue}
Trachea & 17,302 & 46 & 645 & 2 \\
\midrule[0.35pt]
\rowcolor{TableRowGrey}
Spine & 26,426 & 0 & 9,381 & 0 \\
\midrule[0.35pt]
\rowcolor{TableRowBlue}
Right clavicle & 6,765 & 116 & 1,559 & 35 \\
\midrule[0.35pt]
\rowcolor{TableRowGrey}
Left clavicle & 1,534 & 115 & 198 & 34 \\
\midrule[0.35pt]
\rowcolor{TableRowBlue}
Aortic arch & 8,939 & 86 & 2,815 & 30 \\
\midrule[0.35pt]
\rowcolor{TableRowGrey}
Mediastinum & 113,222 & 394 & 26,822 & 97 \\
\midrule[0.35pt]
\rowcolor{TableRowBlue}
Upper mediastinum & 2,820 & 283 & 373 & 77 \\
\midrule[0.35pt]
\rowcolor{TableRowGrey}
SVC & 8,576 & 12 & 1,060 & 2 \\
\midrule[0.35pt]
\rowcolor{TableRowBlue}
Cardiac silhouette & 119,304 & 490 & 29,285 & 113 \\
\midrule[0.35pt]
\rowcolor{TableRowGrey}
Cavoatrial junction & 2,578 & 2 & 334 & 0 \\
\midrule[0.35pt]
\rowcolor{TableRowBlue}
Right atrium & 4,427 & 6 & 946 & 0 \\
\midrule[0.35pt]
\rowcolor{TableRowGrey}
Carina & 6,552 & 32 & 13 & 0 \\
\midrule[0.35pt]
\rowcolor{TableRowBlue}
Abdomen & 25,310 & 0 & 3,348 & 0 \\
\end{longtable}
\endgroup

\FloatBarrier

\begin{table*}[!ht]
\centering
\small
\renewcommand{\arraystretch}{1}
\caption{CheXbert positive-label counts in the training, validation and test splits of MIMIC-CXR and MIMIC-ABN.}
\label{supp-tab:CheXbert_label_counts}
\rowcolors{3}{TableRowBlue}{TableRowGrey}
\setlength{\tabcolsep}{4.5pt}
\begin{tabular}{lrrrrrr}
\toprule
\rowcolor{TableHeader}
& \multicolumn{3}{c}{\textbf{MIMIC-CXR}}
& \multicolumn{3}{c}{\textbf{MIMIC-ABN}} \\
\cmidrule(lr){2-4}\cmidrule(lr){5-7}
\rowcolor{TableHeader}
\multirow{-2}{*}{\textbf{Observation}}
& \multicolumn{1}{c}{\textbf{Train}} & \multicolumn{1}{c}{\textbf{Validation}} & \multicolumn{1}{c}{\textbf{Test}}
& \multicolumn{1}{c}{\textbf{Train}} & \multicolumn{1}{c}{\textbf{Validation}} & \multicolumn{1}{c}{\textbf{Test}} \\
\midrule
Enlarged Cardiomediastinum & 29,449 & 253 & 687 & 2,624 & 27 & 54 \\
\midrule[0.35pt]
Cardiomegaly & 43,528 & 333 & 1,047 & 7,077 & 58 & 131 \\
\midrule[0.35pt]
Lung Opacity & 42,613 & 322 & 949 & 10,854 & 91 & 205 \\
\midrule[0.35pt]
Lung Lesion & 6,577 & 74 & 165 & 2,130 & 25 & 53 \\
\midrule[0.35pt]
Edema & 21,952 & 186 & 638 & 2,488 & 27 & 68 \\
\midrule[0.35pt]
Consolidation & 9,006 & 77 & 228 & 1,360 & 13 & 33 \\
\midrule[0.35pt]
Pneumonia & 14,808 & 117 & 324 & 3,141 & 30 & 63 \\
\midrule[0.35pt]
Atelectasis & 44,209 & 352 & 840 & 8,480 & 71 & 106 \\
\midrule[0.35pt]
Pneumothorax & 5,508 & 37 & 73 & 302 & 6 & 0 \\
\midrule[0.35pt]
Pleural Effusion & 37,209 & 316 & 945 & 4,418 & 36 & 96 \\
\midrule[0.35pt]
Pleural Other & 3,950 & 40 & 93 & 1,300 & 17 & 30 \\
\midrule[0.35pt]
Fracture & 6,374 & 32 & 141 & 1,479 & 10 & 39 \\
\midrule[0.35pt]
Support Devices & 45,398 & 346 & 1,004 & 4,757 & 34 & 102 \\
\midrule[0.35pt]
No Finding & 36,727 & 290 & 153 & 4,664 & 31 & 30 \\
\bottomrule
\end{tabular}
\end{table*}
\FloatBarrier

\begin{table}[!ht]
\centering
\caption{Reasoning length and anatomical grounding of MCoT training trajectories.}
\label{supp-tab:mcot_trajectory_statistics}
\begingroup
\small
\renewcommand{\arraystretch}{1.25}
\setlength{\tabcolsep}{6pt}
\setlength{\extrarowheight}{0.5pt}
\begin{tabular}{>{\raggedright\arraybackslash}m{196pt}>{\raggedleft\arraybackslash}m{45pt}>{\raggedleft\arraybackslash}m{30pt}>{\raggedleft\arraybackslash}m{30pt}}
\toprule[0.8pt]
\rowcolor{TableHeader}
\textbf{Characteristic} & \textbf{Mean} & \textbf{Min} & \textbf{Max} \\
\midrule[0.45pt]
\rowcolor{TableRowBlue}
Reasoning text length (words) & 399.98 & 118 & 823 \\
\rowcolor{TableRowGrey}
Region-grounding instances per trajectory & 5.75 & 0 & 15 \\
\bottomrule[0.8pt]
\end{tabular}
\footnotetext{\footnotesize Statistics summarize $n = 159{,}416$ training trajectories. Word counts include only prose within \texttt{<think>}, after removing complete anatomical grounding blocks (anatomical labels, bounding-box coordinates and \texttt{<image>} markers) and other special markers. MCoT, multimodal chain-of-thought.}
\endgroup
\end{table}
\FloatBarrier

\Needspace{14\baselineskip}
\section{Reasoning CheXbert validation results}
\label{supp-app:reasoning_chexbert}

\begin{table*}[!ht]
\centering
\small
\renewcommand{\arraystretch}{1}
\caption{Reasoning CheXbert performance on the held-out split of the MCoT training set. Uncertain labels are treated as positive for metric computation.}
\label{supp-tab:reasoning_chexbert_metrics}
\begin{tabular}{lccc}
\toprule
\rowcolor{TableHeader}
\textbf{Average} & \textbf{Precision} & \textbf{Recall} & \textbf{$\mathrm{F}_{1}$} \\
\midrule
\rowcolor{TableRowBlue}
Micro & 0.9160 & 0.9194 & 0.9177 \\
\midrule[0.35pt]
\rowcolor{TableRowGrey}
Macro & 0.9110 & 0.9009 & 0.9056 \\
\bottomrule
\end{tabular}
\end{table*}
\FloatBarrier

On this held-out split, reasoning CheXbert achieved micro and macro $\mathrm{F}_{1}$ scores of 0.9177 and 0.9056, respectively (Supplementary Table~\ref{supp-tab:reasoning_chexbert_metrics}). These within-dataset results support its use as an automated label extractor for the reported MCoT analyses.

\Needspace{20\baselineskip}
\section{Model size and inference time}
\label{supp-app:inference_profile}

\FloatBarrier
\begin{table*}[!ht]
\centering
\caption{Model size, decoding strategy and mean inference time on the MIMIC-CXR test set.}
\label{supp-tab:mimic_inference_profile}
\begingroup
\small
\renewcommand{\arraystretch}{1.30}
\setlength{\tabcolsep}{2.4pt}
\setlength{\extrarowheight}{0.5pt}
\setlength{\aboverulesep}{0pt}
\setlength{\belowrulesep}{0pt}
\begin{tabular}{>{\raggedright\arraybackslash}m{79.5pt}>{\centering\arraybackslash}m{54pt}>{\centering\arraybackslash}m{73.5pt}>{\centering\arraybackslash}m{65.5pt}>{\centering\arraybackslash}m{75pt}}
\toprule[0.8pt]
\rowcolor{TableHeader}
\textbf{Model}
& \textbf{Parameters\newline (B)}
& \textbf{Decoding}
& \textbf{Output}
& \textbf{Mean time\newline (s/sample)} \\
\midrule[0.45pt]
\rowcolor{TableRowBlue}
MLRG & 0.30 & Beam search (3) & Report & \makebox[2.5em][r]{0.51} \\
\rowcolor{TableRowGrey}
PriorRG & 0.34 & Beam search (3) & Report & \makebox[2.5em][r]{0.32} \\
\rowcolor{TableRowBlue}
MedGemma-1.5 & 4.30 & Greedy & Report & \makebox[2.5em][r]{1.30} \\
\rowcolor{TableARIVE}
\textbf{ARIVE} & 4.85 & Sampling & MCoT + report & \makebox[2.5em][r]{1.13} \\
\rowcolor{TableRowBlue}
MAIRA-2 & 6.88 & Greedy & Report & \makebox[2.5em][r]{1.20} \\
\rowcolor{TableRowGrey}
Libra & 6.89 & Beam search (10) & Report & \makebox[2.5em][r]{15.03} \\
\bottomrule[0.8pt]
\end{tabular}
\par\vspace{5pt}
\begin{minipage}{\textwidth}
\footnotesize\raggedright
All models were run on a single NVIDIA A100 GPU using the inference protocol specified by each project. Time is averaged per test sample for the complete output shown. B, billion parameters; numbers in parentheses indicate beam width. MCoT, multimodal chain-of-thought.
\end{minipage}
\endgroup
\end{table*}
\FloatBarrier

\clearpage
\section{Prompt templates}
\label{supp-app:prompt_templates}

Supplementary Tables~\ref{supp-tab:prompt_textual_cot} and~\ref{supp-tab:prompt_bbox_insertion} provide the prompts for textual CoT construction and subsequent bounding-box insertion. Supplementary Tables~\ref{supp-tab:prompt_mimic_cxr} and~\ref{supp-tab:prompt_mimic_abn} provide the report-generation prompts for MIMIC-CXR and MIMIC-ABN, respectively. Braced uppercase fields denote placeholders, and {\fontencoding{T1}\ttfamily \textless{}image\textgreater{}} marks an image input. Double square brackets delimit optional prior-study content and are explanatory notation; the enclosed content is included only when a valid prior study is available.

\begingroup
\fontsize{9}{10.5}\selectfont
\setlength{\LTleft}{0pt}
\setlength{\LTright}{0pt}
\setlength{\LTcapwidth}{\textwidth}
\setlength{\LTpre}{4pt}
\setlength{\LTpost}{6pt}
\setlength{\tabcolsep}{8pt}
\renewcommand{\arraystretch}{1.05}

\begin{longtable}{>{\raggedright\arraybackslash}p{\dimexpr\textwidth-2\tabcolsep\relax}}
\caption{Prompt for textual CoT construction. The source report supplies the current findings; few-shot examples specify the narrative style.}\label{supp-tab:prompt_textual_cot}\\
\toprule
\rowcolor{TableHeader}\textbf{Textual CoT construction}\\
\midrule
\endfirsthead
\multicolumn{1}{l}{\tablecaptionfont\textbf{Supplementary Table~\thetable} (continued)}\\
\toprule
\rowcolor{TableHeader}\textbf{Textual CoT construction}\\
\midrule
\endhead
\midrule
\multicolumn{1}{r}{\footnotesize\textit{Continued on next page}}\\
\endfoot
\bottomrule
\endlastfoot
You are an expert chest X-ray radiologist. Produce a complete, optimized first-person reasoning narrative for the current CXR.\\[1.5pt]
The examples define only the desired narrative voice, regional progression, paragraph development, and reasoning depth. Never transfer their clinical findings or routine normal-structure checklist to the current case.\\[1.5pt]
{\fontencoding{T1}\ttfamily \textless{}few\_shot\_examples\textgreater{}}\allowbreak{}\newline {\fontencoding{T1}\ttfamily \{FEW\_SHOT\_EXAMPLES\}}\allowbreak{}\newline {\fontencoding{T1}\ttfamily \textless{}/few\_shot\_examples\textgreater{}}\allowbreak{}\\[1.5pt]
{\fontencoding{T1}\ttfamily \textless{}input\textgreater{}}\allowbreak{}\newline Current CXR report (ground truth; do not copy verbatim):\newline {\fontencoding{T1}\ttfamily \{CURRENT\_CXR\_REPORT\}}\allowbreak{}\\[1.5pt]
Clinical context (available Indication, History, Technique, and Comparison):\newline {\fontencoding{T1}\ttfamily \{CLINICAL\_CONTEXT\}}\allowbreak{}\\[1.5pt]
Prior CXR report:\newline {\fontencoding{T1}\ttfamily \{PRIOR\_CXR\_REPORT\_OR\_NONE\}}\allowbreak{}\newline {\fontencoding{T1}\ttfamily \textless{}/input\textgreater{}}\allowbreak{}\\[1.5pt]
Internally complete two steps:\newline 1. Develop a detailed, region-by-region walkthrough of the current findings, integrating the clinical context and supported prior comparisons.\\
2. Refine the narrative against the supplied evidence, correcting unsupported statements and restoring omitted findings without compressing the reasoning.\\[1.5pt]
Return only the final refined narrative; do not reveal the intermediate draft or audit.\\[1.5pt]
Factual rules:\newline - Use the current report as the authority for current imaging observations.\\
- Preserve every finding, location, measurement, severity, uncertainty, and visibility limitation.\\
- Use clinical context to frame the clinical question and interpretation, without treating history as an observed imaging finding.\\
- Describe reported positive and negative findings. Do not assume that an unreported structure, device, or image-quality feature is normal.\\
- Expand diagnostic reasoning, not visual detail. Blend supported, conditional explanations naturally into the observations.\\
- Retain reported device endpoints and stable contours without adding unsupported placement, function, safety, or normality judgments.\\[1.5pt]
Comparison rules:\newline - Use a valid prior report for supported comparisons of corresponding findings, guided by the supplied Comparison field.\\
- Preserve the certainty of reported changes. A prior-only finding or an omission in either report does not establish a current finding or interval change.\\
- Do not invent dates.\\
- If no valid prior report is supplied, preserve interval findings explicitly stated in the current report without claiming direct prior-image review.\\[1.5pt]
Style and organization:\newline - Match the examples' detailed first-person, reader-guided narrative.\\
- Follow their regional progression, covering technique and limitations, devices, airways, lungs, pleura, central thoracic structures, and other structures only where the case supplies relevant findings.\\
- Develop substantial paragraphs connecting observations, their significance, relevant findings, and the clinical question.\\
- Use supplied technique information and describe the frontal examination only. Never mention lateral views.\\
- End with a concise synthesis of the main findings and clinical implications.\\
- Avoid recommendations and commentary about report wording or missing inputs.\\
- Do not use titles, section headers, lists, bullet points, or bounding boxes.\\[1.5pt]
Output only the complete final narrative.\\
\end{longtable}

\begin{longtable}{>{\raggedright\arraybackslash}p{\dimexpr\textwidth-2\tabcolsep\relax}}
\caption{Prompt for inserting anatomical bounding boxes into textual CoT trajectories to construct MCoT.}\label{supp-tab:prompt_bbox_insertion}\\
\toprule
\rowcolor{TableHeader}\textbf{Bounding-box insertion}\\
\midrule
\endfirsthead
\multicolumn{1}{l}{\tablecaptionfont\textbf{Supplementary Table~\thetable} (continued)}\\
\toprule
\rowcolor{TableHeader}\textbf{Bounding-box insertion}\\
\midrule
\endhead
\midrule
\multicolumn{1}{r}{\footnotesize\textit{Continued on next page}}\\
\endfoot
\bottomrule
\endlastfoot
You are an expert radiologist. Insert relevant bounding boxes from the Anatomical Information into the Generated Reasoning.\\[1.5pt]
Inputs:\newline - Original Report: {\fontencoding{T1}\ttfamily \{ORIGINAL\_REPORT\}}\allowbreak{}\newline - Generated Reasoning: {\fontencoding{T1}\ttfamily \{GENERATED\_REASONING\}}\allowbreak{}\newline - Anatomical Information:\newline {\fontencoding{T1}\ttfamily \{ANATOMICAL\_INFORMATION\}}\allowbreak{}\\[1.5pt]
Rules:\newline - Preserve the original content and wording of the Generated Reasoning.\\
- Insert only bounding boxes whose anatomical regions are explicitly mentioned or directly implied in the reasoning.\\
- Insert each bounding box at most once, immediately before the sentence describing its corresponding region.\\
- Never add boxes for irrelevant or unmentioned regions.\\
- Avoid adjacent bounding boxes; when multiple boxes would be consecutive, retain only the most clinically significant one.\\
- Use exactly this format:\newline   {\fontencoding{T1}\ttfamily\char96}{\fontencoding{T1}\ttfamily \textless{}\textbar{}box\_start\textbar{}\textgreater{}}\allowbreak{} x1 y1 x2 y2 {\fontencoding{T1}\ttfamily \textless{}\textbar{}box\_end\textbar{}\textgreater{}}\allowbreak{}{\fontencoding{T1}\ttfamily\char96}\\[1.5pt]
Return only the updated Generated Reasoning, with no explanations or additional text.\\
\end{longtable}

\clearpage
\begin{longtable}{>{\raggedright\arraybackslash}p{\dimexpr\textwidth-2\tabcolsep\relax}}
\caption{Prompt for MIMIC-CXR report generation, covering the full findings description. The \texttt{[...]} block is included only when an indication is available; \texttt{[[...]]} blocks are included only when a prior study is available.}\label{supp-tab:prompt_mimic_cxr}\\
\toprule
\rowcolor{TableHeader}\textbf{MIMIC-CXR report generation}\\
\midrule
\endfirsthead
\multicolumn{1}{l}{\tablecaptionfont\textbf{Supplementary Table~\thetable} (continued)}\\
\toprule
\rowcolor{TableHeader}\textbf{MIMIC-CXR report generation}\\
\midrule
\endhead
\midrule
\multicolumn{1}{r}{\footnotesize\textit{Continued on next page}}\\
\endfoot
\bottomrule
\endlastfoot
You are an assistant in radiology, responsible for analyzing medical imaging studies and generating detailed, structured, and accurate radiology reports.\\[1.5pt]
[Indication: {\fontencoding{T1}\ttfamily \{INDICATION\}}\allowbreak{}]\\[1.5pt]
\ensuremath{[\![}Prior chest X-ray:\newline {\fontencoding{T1}\ttfamily \textless{}image\textgreater{}}\allowbreak{}\newline Prior findings:{\fontencoding{T1}\ttfamily \{PRIOR\_CXR\_REPORT\}}\allowbreak{}\newline \ensuremath{]\!]}\newline Current chest X-ray:\newline {\fontencoding{T1}\ttfamily \textless{}image\textgreater{}}\allowbreak{}\newline Before answering, think step by step internally. In your internal thinking, output the bounding box of the key diagnostic region in the format {\fontencoding{T1}\ttfamily \textless{}\textbar{}object\_ref\_start\textbar{}\textgreater{}}\allowbreak{}anatomical object name{\fontencoding{T1}\ttfamily \textless{}\textbar{}object\_ref\_end\textbar{}\textgreater{}}\allowbreak{}{\fontencoding{T1}\ttfamily \textless{}\textbar{}box\_start\textbar{}\textgreater{}}\allowbreak{}x1 y1 x2 y2{\fontencoding{T1}\ttfamily \textless{}\textbar{}box\_end\textbar{}\textgreater{}}\allowbreak{}. Generate a comprehensive and detailed description of findings based on the current chest X-ray image and the indication.\ensuremath{[\![} Include a thorough comparison with a prior chest X-ray, emphasizing any significant changes, progression, or improvement.\ensuremath{]\!]}\\
\end{longtable}

\begin{longtable}{>{\raggedright\arraybackslash}p{\dimexpr\textwidth-2\tabcolsep\relax}}
\caption{Prompt for MIMIC-ABN report generation, restricted to abnormal findings. The \texttt{[...]} block is included only when an indication is available; \texttt{[[...]]} blocks are included only when a prior study is available.}\label{supp-tab:prompt_mimic_abn}\\
\toprule
\rowcolor{TableHeader}\textbf{MIMIC-ABN report generation}\\
\midrule
\endfirsthead
\multicolumn{1}{l}{\tablecaptionfont\textbf{Supplementary Table~\thetable} (continued)}\\
\toprule
\rowcolor{TableHeader}\textbf{MIMIC-ABN report generation}\\
\midrule
\endhead
\midrule
\multicolumn{1}{r}{\footnotesize\textit{Continued on next page}}\\
\endfoot
\bottomrule
\endlastfoot
You are an assistant in radiology, responsible for analyzing medical imaging studies and generating detailed, structured, and accurate radiology reports.\\[1.5pt]
[Indication: {\fontencoding{T1}\ttfamily \{INDICATION\}}\allowbreak{}]\\[1.5pt]
\ensuremath{[\![}Prior chest X-ray:\newline {\fontencoding{T1}\ttfamily \textless{}image\textgreater{}}\allowbreak{}\newline Prior abnormal findings:{\fontencoding{T1}\ttfamily \{PRIOR\_ABNORMAL\_FINDINGS\}}\allowbreak{}\newline \ensuremath{]\!]}\newline Current chest X-ray:\newline {\fontencoding{T1}\ttfamily \textless{}image\textgreater{}}\allowbreak{}\newline Before answering, think step by step internally. In your internal thinking, output the bounding box of the key diagnostic region in the format {\fontencoding{T1}\ttfamily \textless{}\textbar{}object\_ref\_start\textbar{}\textgreater{}}\allowbreak{}anatomical object name{\fontencoding{T1}\ttfamily \textless{}\textbar{}object\_ref\_end\textbar{}\textgreater{}}\allowbreak{}{\fontencoding{T1}\ttfamily \textless{}\textbar{}box\_start\textbar{}\textgreater{}}\allowbreak{}x1 y1 x2 y2{\fontencoding{T1}\ttfamily \textless{}\textbar{}box\_end\textbar{}\textgreater{}}\allowbreak{}. Generate a comprehensive and detailed description of abnormal findings only based on the current chest X-ray image and the indication.\ensuremath{[\![} Include a thorough comparison with a prior chest X-ray, emphasizing any significant changes, progression, or improvement.\ensuremath{]\!]}\\
\end{longtable}

\endgroup

\clearpage
\endgroup